\documentclass[jair,twoside,11pt,theapa]{article}
\usepackage{jair, theapa, rawfonts}

\jairheading{?}{2024}{?-?}{10/24}{?/?}
\ShortHeadings{A Divide, Align \& Conquer Strategy For Program Synthesis}
{Witt, Dumančić, Guns, \& Carbon}
\firstpageno{1}

\usepackage{etoolbox}
\apptocmd{\sloppy}{\hbadness 10000\relax}{}{}

\usepackage[dvipsnames]{xcolor}
\usepackage{wrapfig}

\usepackage{stfloats}

\usepackage[export]{adjustbox}

\usepackage{fancyvrb}

\usepackage{stfloats}
\usepackage{multirow}

\usepackage{todonotes}
\usepackage{booktabs}

\usepackage{lstautogobble}

\usepackage{subcaption}

\usepackage{stmaryrd}
\usepackage{listings}
\usepackage{enumitem}
\usepackage{makecell}

\usepackage{graphicx}
\usepackage{fancyvrb}

\usepackage{algorithm}
\usepackage{algpseudocode}
\usepackage{algorithmicx}
\algnewcommand\LeftComment[2]{%
\hspace{#1em}$\triangleright$ \eqparbox{}{#2} \hfill %
}

\usepackage{eqparbox}

\usepackage{verbatim}

\usepackage{amsmath}
\usepackage{amssymb} 
\usepackage{textcomp, gensymb} 
\usepackage{mathtools} 
\usepackage{makecell}
\usepackage{cleveref}
\usepackage{scalerel} 

\newcommand{\ignore}[1]{}

\newcommand\crossshape{\scalerel*{\includegraphics{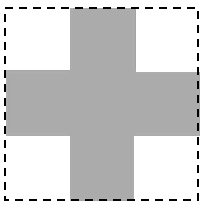}}{X}}
\newcommand\squareshape{\scalerel*{\includegraphics{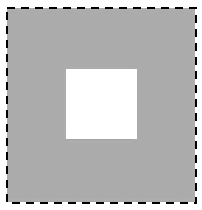}}{X}}

\begin{document}

\title{A Divide, Align \& Conquer Strategy For Program Synthesis}

\author{\name Jonas Witt \email jonas.witt.lab@uni-bamberg.de\\
       \addr University of Bamberg, Markusplatz 3, 96047 Bamberg,\\
       Germany
       \AND
       \name Sebastijan Dumančić \email s.dumancic@tudelft.nl\\
       \addr TU Delft, Van Mourik Broekmanweg 6, 2628 XE Delft,\\
       The Netherlands
       \AND
       \name Tias Guns \email tias.guns@kuleuven.be\\
       \addr KU Leuven, Celestijnenlaan 200a, 3001 Leuven, Belgium
       \AND
       \name Claus-Christian Carbon \email ccc@uni-bamberg.de\\
       \addr University of Bamberg, Markusplatz 3, 96047 Bamberg,\\
       Germany
       }


\maketitle

\begin{abstract}
A major bottleneck in search-based program synthesis, which learns programs from input/output examples, is the synthesis of large programs. As the size of the target program increases so does the search depth which leads to an exponentially growing number of candidate programs.
Humans mitigate the combinatorial explosion that arises from deep program search: they build complex programs from smaller parts.
We introduce a new strategy for program synthesis called Divide, Align \& Conquer (DA\&C) which exploits the compositionality of real world domains to guide the synthesis towards useful sub programs.
\textit{Divide} decomposes each example using a segmentation procedure that is synthesized as part of the learning problem. \textit{Align} matches the components in the decomposed input/output examples in order to steer the search towards combinations which lead to the synthesis of useful sub progams and \textit{Conquer} then solves a standalone synthesis problem on each pair of aligned input/output components.
We show how replacing a deep program search by a linear number of much smaller synthesis tasks leads us to efficiently discover useful sub programs which are then combined into a solution program.
Our agent outperforms current Inductive Logic Programming (ILP) methods on string transformation tasks even with minimal knowledge priors. Unlike existing methods, the predictive accuracy of our agent monotonically increases for additional examples. It approximates an average time complexity of $\mathcal{O}(m)$ in the size $m$ of subprograms for highly structured and, hence decomposable domains such as strings. Finally, we demonstrate the scalability of our technique on high-dimensional abstract visual reasoning tasks from the Abstract Reasoning Corpus (ARC) for which ILP methods were previously infeasible. We are competitive with state-of-the-art agents outside of ILP despite generating only 0.2\% as many candidate programs from a knowledge prior of only 11 generic geometric primitives.
\end{abstract}

\section{Introduction}
A key challenge in program synthesis \cite{gulwani_programming_2017}, which is concerned with learning programs from examples, is the synthesis of large programs.
Program synthesis is often framed as a search problem over a space of programs and, therefore, the larger the program, the more difficult it is to find \cite{alur_search-based_2018}.

\begin{figure}[!ht]
     \centering
     \begin{subfigure}{0.4\linewidth}
        \centering
        \includegraphics[scale=0.26]{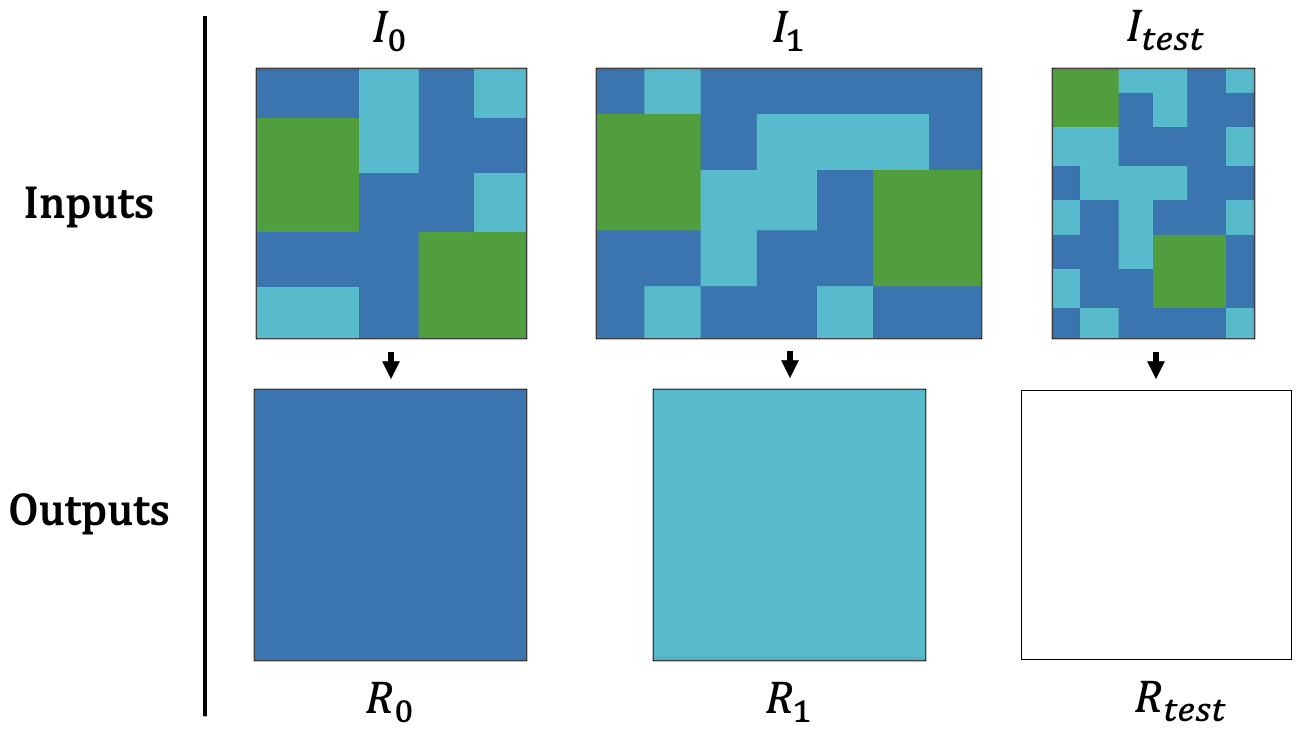}
        \caption{Abstract visual reasoning task.}
        \label{fig:exampletask_ARC}
     \end{subfigure}
     \hfill
     \begin{subfigure}{0.58\linewidth}
        \centering
        \includegraphics[scale=0.34]{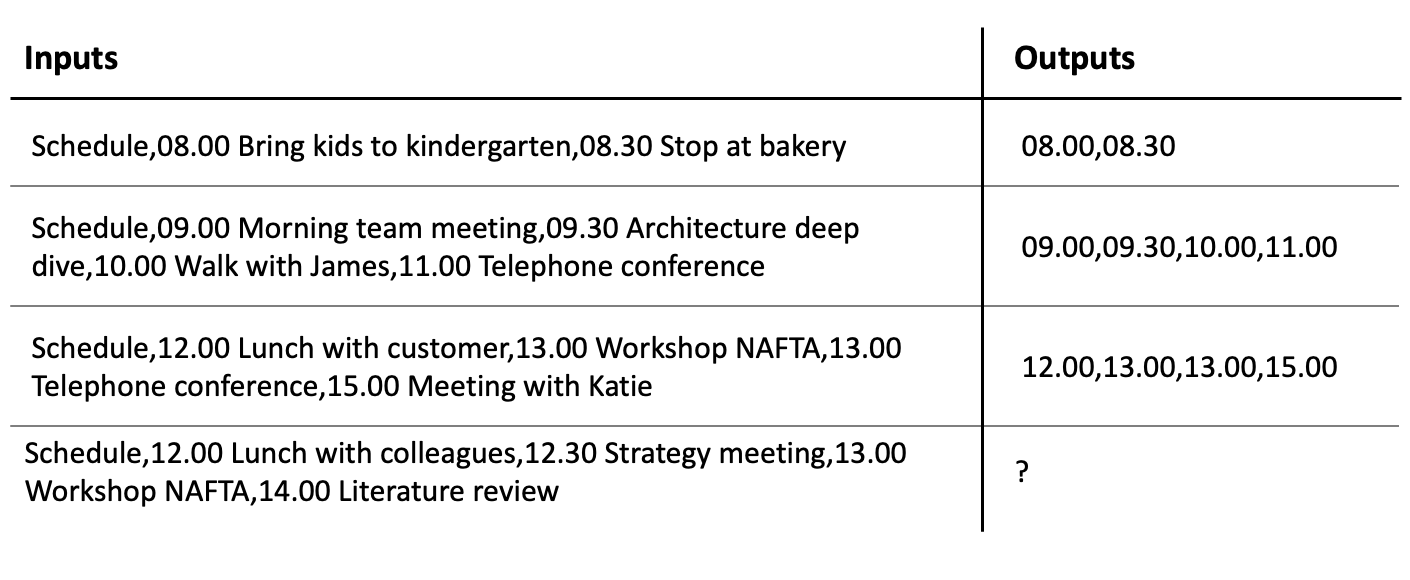}
        \caption{String transformation task.}
        \label{fig:exampletask_strings}
     \end{subfigure}
    \caption{Programming by Examples (PBE) tasks from the Abstraction and Reasoning Corpus (ARC) \cite{chollet_measure_2019} (\Cref{fig:exampletask_ARC}) and the real world string transformations data set \cite{cropper_learning_2020} (\Cref{fig:exampletask_strings}). Agents must search for a program that transforms inputs into outputs. In \Cref{fig:exampletask_ARC}: ``Color the output light-blue whenever there is a light-blue connecting pathway between the green squares in the input.'' In \Cref{fig:exampletask_strings}: ``Extract all times from the meeting schedule and concatenate them using commas.''}
    \label{fig:exampletasks}
\end{figure}


Divide-and-conquer (D\&C) strategies are a common solution to this problem \cite{alur_synthesis_2015}.
Existing D\&C strategies break down the synthesis task by dividing the \textit{set of examples} into subsets \cite{cropper_inductive_2022,alur_scaling_2017,cropper_learning_2022}. 
Each subset of examples defines an independent synthesis task to be \textit{conquered}.
The solutions to the individual synthesis tasks, defined over subsets of examples, are later combined into a global solution i.e. the final program.
Imagine for example a synthesis task in which the input is a pair of numbers $(N1,N2)$ and the desired output is the larger of the two numbers. 
It is easy to find solutions for individual examples, e.g. simply return either $N1$ or $N2$. In order to obtain the task solution program, these sub programs  are combined with an appropriate branching condition ('if $N1 > N2$ then ...').
Given that we discover sub programs independently of the branching condition, we only ever search for expressions of length 1 even though the global solution program has a depth of two (the 'if branch' and its 'if condition'/ 'if consequent').
This is how standard D\&C leads to an exponential decrease in search space through a linear reduction in search depth.

However, standard D\&C makes two assumptions which limit the types of problems on which it achieves a significant decrease in search space: (1) It assumes a solution program can be split into individual pieces which in turn can be identified from non-overlapping subsets of examples. (2) Solutions to subsets of examples are easier to synthesize than a solution covering all examples.
Consider the task in \Cref{fig:exampletask_ARC} in which the color of the output is determined by whether there exists a light-blue path between the green squares in the input image. 
Dividing the task into subsets of examples (with and w/o connecting paths) only marginally simplifies the problem: the search for a program that checks the path's existence remains equally difficult. We extend D\&C to the level of components within a single example, in order to efficiently discover program pieces needed to solve an example. 

\begin{figure}[t]
     \centering
     \begin{subfigure}{0.55\linewidth}
        \centering
        \includegraphics[height=42mm]{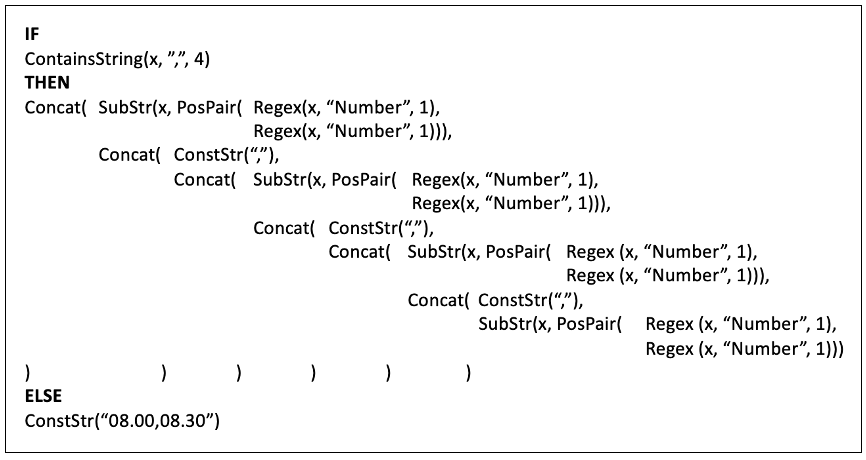}
        \caption{Program synthesized from the PROSE SDK.}
        \label{fig:string_transformation_FlashFill}
     \end{subfigure}
     \begin{subfigure}{0.40\linewidth}
        \centering
        \includegraphics[height=42mm]{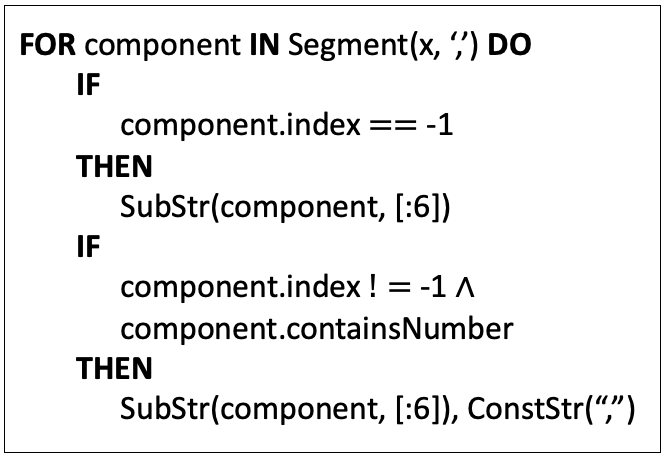}
        \caption{Program synthesized DA\&C (ours).}
        \label{fig:string_transformation_BEN}
     \end{subfigure}
    \caption{A divide, align, \& conquer (DA\&C) strategy yields a compact and well-generalizing program for the task in \Cref{fig:exampletask_strings}.}
    \label{fig:flashfill-BEN}
\end{figure}

In this work, we explore an alternative  \textit{divide}, \textit{align}, \& \textit{conquer} strategy (DA\&C) which divides each example into a set of independent components that are conquered separately. Its atomic operation is to discover a program that solves parts of an example compared to standard D\&C which requires a solution of an example as its atomic operation.
Consider the task in \Cref{fig:exampletask_strings}: It takes a meeting schedule as input and extracts from it a list of times.
A traditional synthesis approach
\cite{gulwani_automating_2011} generates the solution program in \Cref{fig:string_transformation_FlashFill}\footnote{The program was produced using the text transformation API of the Microsoft Program Synthesis using Examples SDK PROSE, a fleet of program synthesis APIs that are the basis of commercial tools like FlashFill in Microsoft Excel. The code is available on GitHub \cite{sumit_gulwani_microsoft_2023}.}.
The program correctly solves the task, however, it is cumbersome and does not generalize to varying lengths of similar formatted meeting schedules.
The task becomes much easier once we acknowledge that examples have inherent compositional structure.
A DA\&C strategy breaks down the inputs/outputs into substrings using commas as delimiters and from those extracts times as prefixes. It flexibly generalizes to schedules of varying lengths (\Cref{fig:string_transformation_BEN}).

Two major challenges in working with smaller components beyond the level of examples are, first, the question of how to \textit{segment} examples into components and, second, the problem of finding meaningful alignments between components in the inputs/outputs.
An alignment between an input and output component is meaningful whenever it leads to a program that reconstructs the component in the output given the component in the input.
An example output is solved if all of its components are successfully reconstructed from components in the input.
In the example of \Cref{fig:exampletask_strings}, synthesis is performed on the aligned components '08.30 Stop at bakery' and '08.30' which leads to the program SubStr({\it component},[:6]).
One could explore every possible correspondence between input/output components but that is likely to diminish the benefit of problem decomposition as the number of synthesis steps (needed to discover that all but one correspondence is meaningless) will be large.
The problem is further complicated by the fact that not every input component needs to be present in the output and, thus, does not need to have a correspondence (e.g. 'Schedule' in \Cref{fig:exampletask_strings}). 

We propose to mitigate this problem by {\it structurally aligning} the input and output scenes, in a process that mimics {\it analogical reasoning}.
A meaningful alignment between two scenes maximizes their shared structure.
For instance, in the last training example in \Cref{fig:exampletask_strings}, there are two conflicting meetings booked at the same time. Any one of them could produce the '13.00,13.00' substring in the output by repetition.
However, only an alignment that respects the ordering of components and aligns '13.00 Workshop NAFTA' with the first '13.00' output substring and '13.00 Telephone conference' with the second '13.00' output substring will produce a minimal solution program that generalizes to arbritary meeting schedules.
In order to arrive at this mapping, we leverage analogy engines such as the structure-mapping engine (SME) \cite{falkenhainer_structure-mapping_1989}, a computational model implementing structure-mapping theory (SMT) \cite{gentner_structure-mapping_1983}, a formal account of analogy making in humans.

We combine the synthesis performed on subsets of components with hierarchical search \cite{wang_synthesizing_2017} which allows us to learn parts of a final solution program sequentially, thereby, efficiently pruning irrelevant areas of the search space: First, segmentation operations are independently learned of the transformation program. The information available about the segmented output components is used to prune transformation programs that do not need to be explored during synthesis (e.g. during the synthesis of a lowercase output component, all programs with 'capitalize' operators can be pruned away). Second, only after we have discovered a successful program (on a tuple of input/output components), we search for similar pairs and learn an abstract classifier which tells us when to apply it (on which components).

In addition, our approach demonstrates progress on two key topics in program synthesis: First, we apply DA\&C to high-dimensional examples without manual pre-processing them into symbolic encodings. The segmentation of examples is discovered as part of the synthesis in order to optimize reconstruction accuracy in the output. This strategy prepares synthesis on raw real-world inputs.
Second, synthesis tools struggle with large knowledge libraries. DA\&C performs synthesis on subsets of input/output components, where the information on the output component is used to prune not applicable background knowledge. This is similar to a relevance mechanism which selects predicates to be used in the current search from a (growing) global library, e.g. the synthesis of a green output object does not need any other color constants besides green.\\
To summarize, we propose a 'divide, align \& conquer' strategy (DA\&C) which performs hierarchical search on partial examples in order to infer generative programs from a minimal number of examples. Specifically, our contributions are as follows:

\begin{enumerate}
    \item We re-conceptualize the D\&C strategy such that synthesis can exploit repeating compositional structure within individual examples. We demonstrate how this allows agents to synthesize more complex programs even on high-dimensional inputs such as bitmaps.
    \item We introduce analogical reasoning as a means to mitigate the combinatorial explosion of correspondences between input/output components. We demonstrate how the information on an aligned output component can be used to prune parts of the program search.
    \item We implement DA\&C in an algorithmic agent called BEN. Its performance is evaluated on the established setting of string transformation tasks and the challenging abstract visual reasoning data sets from the Abstraction and Reasoning Corpus (ARC). \cite{chollet_measure_2019}.
\end{enumerate}

\clearpage


\section{Related Work And Background}

Our work lies in search-based program synthesis, specifically the \textit{Programming by Examples} (PBE) systems which learn programs that are consistent with a semantic specification implicitly defined by a set of input/output examples \cite{gulwani_programming_2017}.

Recent work has focused on two areas: (1) improving the efficiency of search by guiding it towards more promising candidates and (2) rewriting of task specifications to simplify the search problem itself. Our work falls into the second category. It also shares common goals with works in the first area.

\paragraph{Improving search efficiency.} The first area has recently focused on search optimizations that leverage the intermediate states of programs and evaluate their distance to the target solution \cite{ellis_write_2019,cropper_learning_2020,nye_representing_2021}. Often these approaches are neurosymbolic hybrids that statistically learn a heuristics function from past synthesis tasks. DA\&C in its current form does not make use of statistical learning across tasks. Instead, it guides the synthesis search by pruning the search space. It uses the information about an output component (gained from an example segmentation) to prune away irrelevant parts of the search space.
Apart form heuristic search, other approaches use re-representation techniques to condense programs and, thus, make it is easier to search them: compression-driven rewriting or functional abstraction (e.g. predicate invention) \cite{henderson_automatic_2014,cropper_learning_2020-1,dumancic_knowledge_2021}. The programs learned through DA\&C are similarly compact but are produced by \textit{rewriting} the examples, not the program. The way that DA\&C rewrites examples is by searching for a meaningful segmentation into components.
There exist search optimizations which already make use of decomposition operators in order to guide the synthesis: Symbolic backpropagation \cite{gulwani_programming_2017} is a top-down deductive search that uses inverse operators to propagate constraints on the overall solution program to sub-expressions of the intermediate program.
These sub-specifications are used to filter substrings from the input with the help of regular expressions $pos(x,R1,R2,k)$.
This approach is different from DA\&C in that it first assumes an intermediate program and then deductively finds the most probable segmentation to support the program. In contrast, we first pick a segmentation and then inductively search for a program.
The downside of symbolic backpropagation is that it requires inverse operators of each language primitive to work from the example output backwards. \citeA{gulwani_programming_2017} address the combinatorial challenges of inverse synthesis using forwardprop filtering of inverse candidates.
However, the more expressive the transformation language, the larger the number of conceivable inverses. For example, a 'replace' operator which replaces a substring in the input with a constant substring in the output will generate an intractable number of inverses. In DA\&C, 'replace' operators are cheap cause synthesis is executed on pairs of input/output components where the constant substring is available through the output component and its inverse is simply the input component.

\paragraph{Rewriting the search problem.} Previous work in this area has demonstrated D\&C (divide \& conquer) strategies on the level of examples \cite{alur_synthesis_2015}. For instance, \citeA{cropper_learning_2022} combines D\&C with modern constraint solving using answer sets. Successful intermediate programs are used to search for a more general program that applies to multiple chunks until all positive examples are covered. These approaches only work on subsets of examples (called \textit{chunks}), while DA\&C performs synthesis on subsets of components within an example. We use antiunification in the underlying domain-specific language (DSL) to learn branching conditions that combine independent programs on components into a global solution program. Our approach is more similar to \citeA{alur_scaling_2017} who combines intermediate programs using a conditional expression grammar in a multi-label decision tree learning paradigm. However, their approach also only works with subsets of examples.
DA\&C exploits the innate structure of input/output examples to decompose a semantic specification into sub-specifications that are solved in multiple smaller synthesis tasks. This trades some of the exponential complexity of a deep program search with a linear number of additional synthesis tasks.

\paragraph{Analogical reasoning.} In order to mitigate the combinatorial explosion that arises from decomposing examples into components and searching for alignments between components, we make use of {\it analogical reasoning} \cite{evans_heuristic_1964,mitchell_analogy-making_1993}.
Research in the cognitive sciences has highlighted the importance of analogies for problem-solving \cite{hofstadter_analogy_2001,mitchell_abstraction_2021}. In the past, these models were applied to study psychometric tests of intelligence \cite{snow_topography_1984}, e.g. in number series completion, string transformations, verbal analogies, and Raven's matrices.
For instance, \citeA{lovett_modeling_2017} use analogical reasoning to compute 'patterns of variance' (descriptive statements of how objects change) across subsequent scenes within each row of a Raven's matrix. In contrast, we explicitly learn \textit{actionable} transformation programs which are capable of \textit{generating} new outputs.
We make use of the structure-mapping-engine (SME) \cite{falkenhainer_structure-mapping_1989,gentner_structure-mapping_1983}, a computational model of analogical reasoning in humans, to determine how an input segmentation is analogous to the segmented output and derive from it pairwise correspondences. SME is a symbolic approach to structure-mapping which purely relies on the syntactic representation of a problem. It is a good fit for our goal because it works across domains and does not require training on large data sets.

Program synthesis on high-dimensional examples (e.g. bitmaps) has seen much less work than the established domains such as bit vector and string manipulations or generation of invariants \cite{alur_search-based_2018}. \citeA{ellis_unsupervised_2015} have synthesized programs to represent visual concepts and perform item classification using probabilities in a generative process. The synthesis itself was not directly performed on bitmaps but on automatically parsed symbolic encodings. \citeA{cropper_learning_2020} have learned programs to generate bitmaps using an example-dependent loss function instead of logical entailment in order to better guide the synthesis on large programs.
We apply DA\&C to the Abstraction and Reasoning Corpus (ARC) \cite{chollet_measure_2019}, a much more diverse collection of bitmap data sets, that was introduced to foster research on the efficiency with which agents acquire new skills. The corpus is a collection of heterogeneous visual reasoning tasks. On each task, the agent synthesizes a program that takes a bitmap as input in order to generate a bitmap as output (example task in \Cref{fig:exampletask_ARC}). ARC is especially interesting to our work as visual reasoning programs tend to be large (in current program synthesis context) and, thus, out of reach for existing synthesis techniques. Visual scenes also intuitively demonstrate the idea of being composed of objects \cite{johnson_fast_2021,wagemans_century_2012}.
ARC is a challenging benchmark where the top-ranked agents only solve about 20\% of tasks on a hidden test set and perform a brute force 'generate \& test' approach using elaborately handcrafted domain-specific languages (DSL) \cite{wind_1st_2020,de_miquel_2nd_2020}. Instead, we are investigating a systematic program synthesis approach that scales to high-dimensional examples using the idea of task decomposition.

\subsection{Background}
\label{sec:background}

In this subsection, we briefly introduce the idea of program synthesis as search and the structure-mapping theory (SMT) \cite{gentner_structure-mapping_1983} both of which are integral parts of DA\&C. Readers already familiar with these concepts are free to skip ahead.

\paragraph{Program synthesis as search.} Learning a program is formulated as search through a language space. In addition to a semantic specification (e.g. a set of input/output examples), the agent is provided syntactic constraints (e.g. grammar $\mathcal{G}$) on the set of program candidates \cite{alur_search-based_2018}. Every derivation from $G$ is a candidate program. A grammar $\mathcal{G}$ with a finite number of production rules can produce infinitely many programs of increasing length. One of the central challenges in the field is scaling the synthesis to large programs. The longer a solution program, the larger its search space which exponentially grows ($b^{|d|}$) in the depth of program $d$ and the average branching factor $b$ of the grammar. DA\&C contributes to the goal of scaling search-based program synthesis by trading a deep program search with a linear number of smaller synthesis tasks.

\paragraph{Structure-mapping theory.}
\label{par:structuremapping}
Analogies help us reason about an unfamiliar target domain (e.g. the structure of an atom: the relationship between electrons and its nucleus) using existing knowledge of a familiar base domain (e.g. the structure of our solar system: planets orbitting around the sun). Structure-mapping theory (SMT) systematizes the process by which humans perform analogical reasoning. Computational models that implement SMT consume propositional scene representations of a base and target and search for an alignment between the two which maximizes their shared relational structure (systematicity principle): for example, electrons orbit around the nucleus of an atom similar to how planets orbit around the sun. The alignment between the base and target (called mapping) consists of a set of matched entities and predicates, e.g. an electron is to the nucleus what a planet is to the sun because both share the 
relationship of orbitting around an object with greater mass. Mappings are evaluated systematically through three steps:

\begin{enumerate}
    \item Generation of \textit{local match hypotheses} $mh$: Each predicate pair in the base/target is evaluated through a set of match constructor rules. If successful, the predicate pair forms a local match hypothesis. Local match hypotheses will be combined in the following steps to find isomorphic subgraphs between the base and target. Match constructor rules put syntactic contraints on the types of local match hypotheses that are formed. An example of such a constructor rule is given below: In this case, any two predicates with matching functors that are not attributes form a $mh$.
    
    \begin{Verbatim}[fontsize=\small]
(mhc-rule (:filter ?b ?t :test (and (equal (exp-functor ?b) (exp-functor ?t))
            (not (attribute? (exp-functor ?b))))) (construct-mh ?b ?t))
    \end{Verbatim}
    
    In the example above, the constructor rule leads to a match hypothesis being formed over the predicate pairs \textit{(orbits electron nucleus)} and \textit{(orbits planet sun)}. Let's assume that this step also returns a match between the predicate pair \textit{(greater\_mass sun planet)} and \textit{(greater\_mass nucleus electron)}.

    \item Derivation of \textit{global mappings} (GMAPs): A GMAP is a maximal and structurally consistent set of local match hypotheses in the base/target. A GMAP is structurally consistent if (1) all matched predicates also form match hypotheses between their arguments and (2) all $mhs$ in the GMAP yield a consistent set of one-to-one correspondences between entities in the base/target.

    In the example, the predicate pair 'orbits' is structurally consistent if its arguments 'electron'/ 'planet' and 'nucleus'/ 'sun' also form match hypotheses. The set of predicates 'orbits' and 'greater\_mass' is then also structurally consistent because the 'greater\_mass' predicate enforces the same one-to-one entity mapping as 'orbits': 'electron' is paired with 'planet', 'nucleus' is paired with 'sun'.
    
    \item Ranking of GMAPs: All maximal structurally consistent GMAPs are ranked according to their extent of matched relational structure. Nested relational structures are favored over shallow matches which is characteristic of meaningful analogies.

    A meaningful analogy in the running example is the combined set of both the 'orbits' and 'greater\_mass' predicates incl. all derived entity mappings.
\end{enumerate}

We will make use of SMT in order to answer the question: How is the input of an example structurally similar to its output? From there, we systematically search for programs that transform a part of an input into its corresponding (SMT-derived) part in the output. The search will first explore those pairwise correspondences that contribute most to a structurally consistent and maximal alignment between the input/output.

\section{Problem Definition}

More formally, we solve a standard synthesis task represented as a tuple $(\Phi,\mathcal{G})$ of a specification $\Phi$ and a grammar $\mathcal{G}$. The specification is given in the form of (positive) examples ${\cal Q}$, where each example $q_l$ is an input/output pair $q_l = (I_l, R_l)$. A program $p$ is said to solve the task when $p \in \mathcal{G}$ such that ${\forall (I,R) \in {\cal Q}, \; p(I) = R}$.

\begin{figure}[t]
\centering
\includegraphics[scale=0.42]{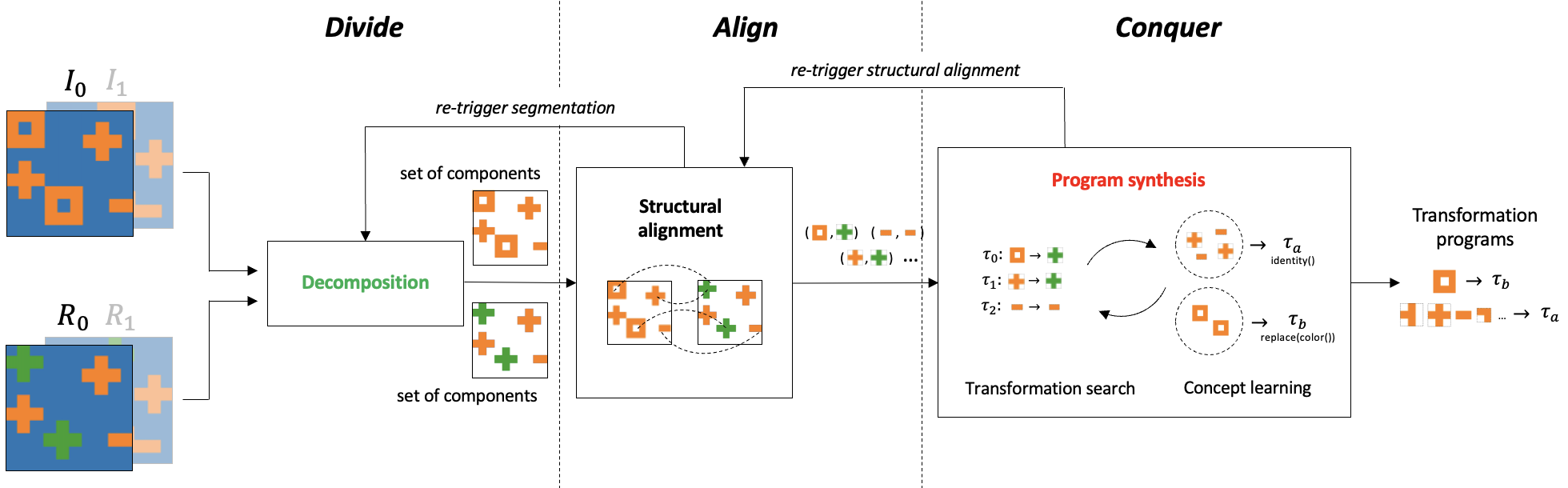}
\caption{Every example is decomposed into component parts (e.g. visual objects in a scene, words in a sentence). We use the information about components and their relations to produce a structural alignment between the input/output scenes. Each correspondence from this alignment is solved independently using off-the-shelf synthesis techniques. Correspondences are considered in the order in which they contribute to the structural alignment (\textit{analogy}). For each unique partial program, we learn a formula which specifies its corresponding input space (the space of components on which it should be executed). The combination of an input space and its transformation is called a transformation rule. A solution program consists of a set of transformation rules that if applied to the components in the input reconstruct all components in the output.}
\label{fig:architecture}
\end{figure}

In this work, we focus on problem domains with separable specifications which is a commonly studied field of synthesis problems \cite{neider_synthesizing_2016}. A specification is separable if it only relates an input to its output and gives no further constraints on the relation between outputs of different inputs, $p(I) = R \; \land \; \Phi(I,R)$ with program $p$ and formula $\Phi$ \cite{alur_synthesis_2015}. All synthesis tasks which make use of examples fulfill this condition because $\Phi$ is implicitly specified over input/output pairs $(I,R)$. We split the problem of synthesizing a program into three subtasks (\Cref{fig:architecture}): \textit{divide}, \textit{align}, and \textit{conquer}.

First, we learn to {\it divide} the input/output examples into sets of components. The purpose of divide is to identify components that repeat across examples. For instance, divide on the task in \Cref{fig:architecture} yields a set of orange and green shapes for each example input/output.

Second, we search for an {\it alignment} of the components in the input set of an example with the components in its output set. The goal of the align step is to identify pairs of input/output components that can be transformed into each other with a minimal program. In the example, the mapping between orange squares in the input and green crosses in the output has high probability because they share the same relational placement within the image. The information on pairwise correspondences is used as a heuristic to guide the search for a successful program. This is why it is no concern that orange crosses in the input are also matched with green crosses in the output due to their surface similarity. These matches are assigned a lower score and are never explored because by that time the search will have already found a set of transformation programs that fully reconstructs all output examples.

Third, we {\it conquer} each pair of aligned input/output components separately which means synthesising a transformation program that reconstructs the output component using the aligned component in the input. In order to combine transformation programs into a global solution program, we learn a concept over all those input components (across examples) which made use of the same transformation.

New test examples are evaluated by first decomposing their examples into components and then evaluating each component against the concepts learned for the transformation programs. If the evaluation succeeds, the transformation is applied to the input component; its result is pasted to the output.

If \textit{align} does not find a meaningful alignment for any of the components in the output, it re-triggers divide to produce a new segmentation of the examples into updated sets of components.
If \textit{conquer} is unable to synthesize a transformation for any of the output components, it re-triggers align to find a new component pair for the missing output component.\\
We now introduce the problem starting from the language space of programs and then formalize the DA\&C paradigm on top of this definition. As is standard in PBE, DA\&C uses syntactic priors to restrict the search space of programs. What distinguishes DA\&C from other PBE methods is that it requires the syntactic priors are assigned to one of the three phases of the hierarchical search (divide, align, conquer). Specifically, DA\&C expects as input constraints on the segmentation of examples $\mathcal{G}_{decomp}$ ('divide'), an encoding scheme for segmented components $\mathcal{G}_{concept}$ ('align'), and a grammar for the transformation programs $\mathcal{G}_{transform}$ ('conquer'). We later show that the total amount of syntactic priors provided to a DA\&C agent across these three inputs is not different from the syntactic priors provided to other synthesis agents within a single synthesis grammar. The process we used to define these inputs for the experimentation domains in this paper (abstract visual reasoning and string transformations) was analogous to how one would define a single grammar: We defined decomposition constraints, component attributes, and transformation primitives, tested, and extended them in a process of iterative refinement. In order to assess the effectiveness of splitting the synthesis grammar across three inputs and probing the dependency of this architecture on custom-engineered domain specific primitives, we later describe an experiment in which we purposefully introduce nonsense primitives into the agent's knowledge base. We also present an extensive ablation analysis in the experiments section.

\begin{figure}[t]
    \caption{Divide-align-conquer synthesis grammar $\mathcal{G}$.}
    \label{fig:problemdefinition}

    \begin{tabular}{ | >{\raggedleft}p{0.2\textwidth} p{0.05\textwidth} p{0.665\textwidth} | }
        \hline
        S & $\rightarrow$ & \textcolor{Green}{DECOMPOSE}, $\mathcal{T}$\\
        \multicolumn{3}{|l|}{}\\
        $\mathcal{T}$ & $\rightarrow$ & $\tau$, $\mathcal{T}$ $|$ $\epsilon$\\[0.05cm]
        $\tau$ & $\rightarrow$ & if \textcolor{Blue}{CONCEPT} then \textcolor{Red}{TRANSFORM}\\
        \multicolumn{3}{|l|}{}\\
        TRANSFORM & $\rightarrow$ & \makecell[tl]{$o_i$ $|$ primitive01(TRANSFORM, args) $|$\\ primitive02(TRANSFORM, args) $|$ ...}\\
        \hline
    \end{tabular}
\end{figure}

\paragraph{Space of programs.} $\mathcal{G}$ is a typed grammar that yields programs $p: \mathcal{B} \rightarrow \mathcal{B}$ which consume and produce a domain-dependent standard type $\mathcal{B}$ (e.g. 2D bitmap, list of chars). Every $p\in \mathcal{G}$ is a composite program with three learnable subroutines (\Cref{fig:problemdefinition}):

\begin{enumerate}
    \item A \textcolor{Green}{decomposition function} $\delta: \mathcal{B} \rightarrow \{\mathcal{B}\}_{i=0}^{N}$ which decomposes an input type $\mathcal{B}$ into a set of components of the same type $\forall q_l: \delta(I_l,R_l) = (\{o_i\}_{i=0}^{N}, \{o_j\}_{j=0}^{M})$. 'Divide' expects a configurable family of decomposition functions (e.g. in the form of a simple set definition or a domain-specific grammar $\mathcal{G}_{decomp}$). It reflects basic syntactic knowledge priors on how components comprise a domain example (e.g. words are separated using whitespaces, a visual scene can be comprised of multiple objects and a partially occluded background).
    We will provide detailed examples in the following sections. In the task of \Cref{fig:architecture}, the learned decomposition function segments example images into objects whose pixels are equally colored and directly neighboring.

    The two remaining subroutines belong to the body $\mathcal{T}$ of the program. They make up the if-statements defined by the synthesis grammar $\mathcal{G}$ in \Cref{fig:problemdefinition}.
    Each if-statement is a transformation rule $\tau: \mathcal{B} \rightarrow \{\mathcal{B}\}_{j=0}^{Y}$ that consumes exactly one component from the input and produces a set of components in the output.
    We explicitly define solution programs using a grammar with top-level branches to make it easy to combine transformations (\Cref{fig:problemdefinition}).

    \item The \textcolor{Red}{transformation program} $p'(o_i) = o_j$ is the result of a synthesis performed on a component tuple $(o_i,o_j)$ using function primitives (e.g. \textit{drop\_first\_char()}, \textit{color(c)}) provided by a domain grammar $\mathcal{G}_{transform}$.
    The task in \Cref{fig:architecture} yields two transformations: One replaces objects of shape \squareshape{} with green objects of shape \crossshape{}. The other copies the remaining objects to the output.
    
    \item The \textcolor{Blue}{rule condition} learns the context in which to apply a transformation.
    This information is used to combine transformations on specific components into a solution program on the task level.
    The rule condition is expressed as a formula over component attributes (e.g. length of a  substring, color of an object) taken from a domain grammar $\mathcal{G}_{concept}$. It is a binary classifier that returns True iff the component belongs to its concept: $\{\mathcal{B}\}_i \rightarrow [\top,\bot]$.
    Because examples demonstrate a recurring logic, there will be component tuples (across examples) that are solved by the same transformation program $p'$, their set is denoted as $CORR_{p'}$. We learn a concept $C_{p'}$ over the input components of tuples in $CORR_{p'}$.

    \begin{Verbatim}[fontsize=\small, commandchars=\\\{\}]
if o.shape == \squareshape then o.replace_by(\crossshape).color(green)
if o.shape != \squareshape then o.identity()
    \end{Verbatim}
        
    The unite operator $\oplus$ combines tuples consisting of a program and its concept $(p'_0,C_{p'_0})$ $(p'_1,C_{p'_1})$ into a single program.

    \begin{equation}
    \label{eq:unificationoperator}
    \tau_0 \oplus \tau_1 = (C_{p'_0} \cdot p'_0) \oplus (C_{p'_1} \cdot p'_1) = \text{ if } C_{p'_0} \text{ then } p'_0 \text{, if } C_{p'_1} \text{ then } p'_1 
    \end{equation}

\end{enumerate}

At test time, the learned decomposition function is applied to each test example and yields a set of components. On each input component $o_i$, we check if it belongs to any of the concepts $C_{p'}$ in the transformation rules and if it does, execute $p'$ to produce a component in the output: $\tau = \text{ iff } o_i \in C_{p'} \text{ then } p'(o_i)$.\\
To summarize, a DA\&C solution assigns to every component in the output $o_j \in \delta(R_l)$ a component $o_i$ in the input $o_i \in \delta(I_l)$,
with $corr(o_i, o_j)$ and a transformation program $\tau_k = C_{p'} \cdot p'$ transforming $o_i$ into $o_j: p'(o_i) = o_j$
such that if all transformation programs $\tau_k \in \mathcal{T}$ are applied to the decomposed input $\delta(I_l) = \{o_i\}_{i=0}^{N}$,
their resulting components fully reconstruct the correct output,
$compose(\bigcup\limits_{i=0}^n\bigcup\limits_{\tau_k}^{\mathcal{T}}\tau_k(o_i)) = R_l$.

The divide, align, and conquer subroutines are interdependent. In the next section, we introduce a specific implementation that leverages these interdependencies: We show how to efficiently discover decomposition functions and correspondences between components that minimize the complexity of a solution program.

\section{Architectural Overview}
We now detail the algorithmic approach to each of the DA\&C parts following the running example of \Cref{fig:architecture} already introduced in the last section.

The goal of the decomposition phase is to segment examples into sets of \textit{meaningful} components.
A set of input components is meaningful whenever it leads the program search to discover compact transformation programs $\mathcal{T}$ that fully reconstruct the output. This idea is captured in the joint probability $P(\mathcal{Q},\delta,\mathcal{T})$. We minimize its negative log-likelihood in order to find a decomposition function $\delta$ and learn a minimal set of transformation rules $\mathcal{T}$ which fulfill the specification given by inputs/outputs $\mathcal{Q}=(I_l,O_l)$.
The joint probability is the result of a generative process which starts from the input examples and applies a decomposition function together with a set of transformation programs to generate output examples.

\begin{equation}
\label{equation:objective}
-log(P(\mathcal{Q},\delta,\mathcal{T})) = -log P(\delta) - \sum_{l=0}^{N}{(log P(R_l|\mathcal{T},\delta(I_l)) + log P(\mathcal{T}|\delta(I_l,R_l))))}
\end{equation}

The terms in \Cref{equation:objective} from left to right are a prior probability on the decomposition function, a likelihood expressed as an example dependent reconstruction accuracy, and a prior on the set of transformation programs. The prior probability of a set of transformation programs is dependent on the components produced by $\delta$ on a specific example $(I_l,R_l)$ because their information is used to prune the space of programs, for example: to reconstruct the green cross in the output (\Cref{fig:architecture}), any program that makes use of a color terminal other than green is pruned away. In contrary, the prior on $\delta$ is moved outside the sum, because it is applied to all examples of a task equally.

\begin{algorithm}[ht]
\caption{Overview Divide-Align \& Conquer}
\label{alg:divide-align-conquer_overview}

\textbf{Input:} specification $\Phi = \{(I, R)\}_{l=0}^{L}$, program grammar: $\mathcal{G}_{decomp}$, $\mathcal{G}_{concept}$, $\mathcal{G}_{transform}$\\
\textbf{Parameters:} search depth $d$ \\
\textbf{Output:} $p\in \mathcal{G}$ s.t. $\forall (I,R)\in \Phi: p(I)=R$
\begin{algorithmic}[1]

\State $\Delta = ExtractAndRank(\mathcal{G}_{decomp})$

\While {$|\Delta| > 0$}
\State $decomposition\_func \leftarrow Pop(\Delta)$
\State $O \leftarrow \bigcup_{l=0}^{L}{decomposition\_func(R_l)}$
\Comment{\textbf{Divide each example}}
\State {$\mathcal{T}' \leftarrow \{\}$, $\mathcal{C} \leftarrow \{\}$}

\While {$|O| \geq 1$}
\State {$o_{curr}^{l} \leftarrow Choose(O)$}
\Statex \LeftComment{4.5} {$l$ is the example index which contains the current output component}
\If {$o_{curr}^{l}$ not previously explored}
\State {$C[o_{curr}^{l}] \leftarrow RankCorrespondences(decomposition\_func(I_l), o_{curr}^{l})$}
\Comment{\textbf{Align}}
\EndIf
\State {$o_i \leftarrow Pop(C[o_{curr}^{l}])$}
\State {$\tau = LearnTransformation((o_i, o_{curr}^{l}), \mathcal{G}_{transform}, \mathcal{G}_{concept}, d)$}
\Comment{\textbf{Conquer}}
\State {$\mathcal{T}' \leftarrow \mathcal{T}' \cup \tau$}

\If {$\forall o_j\in \{decomposition\_func(R_l)_{l=0}^{L}\}$ covered by $\mathcal{T}'$}
\State {\Return {$decomposition\_func, MinimalRuleSet(\mathcal{T}', \delta, \Phi)$}}
\EndIf

\If {$|C[o_{curr}^{l}]|$ == 0}
\State {$O \leftarrow O\setminus o_{curr}^{l}$}
\EndIf

\EndWhile

\EndWhile

\end{algorithmic}
\end{algorithm}

The search is a generate \& test approach outlined in \Cref{alg:divide-align-conquer_overview}. For now, we only consider the high level control flow without optimizations to illustrate its main ideas.
First, we compute an ordered list of decomposition functions $\Delta$ (line 1), ranked in the order of estimated usefulness to the downstream synthesis. The subsequent DA\&C loop proceeds from the highest ranking decomposition function.
The frontier set $O$ keeps track of all segmented components in the example outputs which have not been considered for synthesis. We pick one (line 7) and do greedy best-first search over its ranked list of input components. We learn a transformation program (line 12) on the highest ranked correspondence and add it to the library $\mathcal{T}'$. Every time the library changes, we check if it contains sufficient transformation programs to reconstruct the output components in all examples (line 14) and if it does, return a global solution program that consists of a minimal set of transformation rules. A minimal program set is one that minimizes the optimization function in \Cref{equation:objective}. The current output component $o_{curr}$ is deleted from the frontier set $O$ if all of its correspondences in the input have been explored and the DA\&C loop repeats.

\subsection{\textit{Divide} - Decomposition}
\label{sec:segment}

We learn a bottom-up parse that extracts structured components from examples instead of performing inductive synthesis on unstructured high-dimensional inputs at the level of their atoms (e.g. pixels, letters). 
This is done in the \textit{divide} stage: We synthesize a decomposition function $\delta$ and segment the task examples into complete sets of mutually exclusive components.
In the following sections, we will continue to use abstract visual reasoning as an illustration example. We refer to our agent implementing DA\&C as 'BEN'.

\subsubsection{Segmentation}
\label{sec:segmentation}
The decomposition of a standard type (e.g. string, image) $\delta(\mathcal{B}) \rightarrow \{\mathcal{B}\}_n$ yields a set of components of the same type, called segmentation of $\mathcal{B}$. In this paper, we use a symbolic definition of decomposition functions $\delta$ over a set of constraints $C$ that determine if two atoms (e.g. pixels, characters) are part of the same component.
\textit{Mono-colored} is an example of such a constraint which applies to all objects in the task of \Cref{fig:architecture}. The reason we use a symbolic implementation instead of a non-parametric statistical model (e.g. deep autoencoders, CNNs) is because the domains we focus on only provide a single digit number of training examples, exhibit low noise, and offer intuitive segmentation heuristics that can be readily articulated by human task solvers (e.g. words are separated by whitespaces) \cite{raza_automated_2017}. ARC images also have at most 10 colors and are bounded in size (30x30 pixels).

\begin{figure}[t]
     \centering
     \begin{subfigure}{0.48\textwidth}
        \centering
        \includegraphics[scale=0.2]{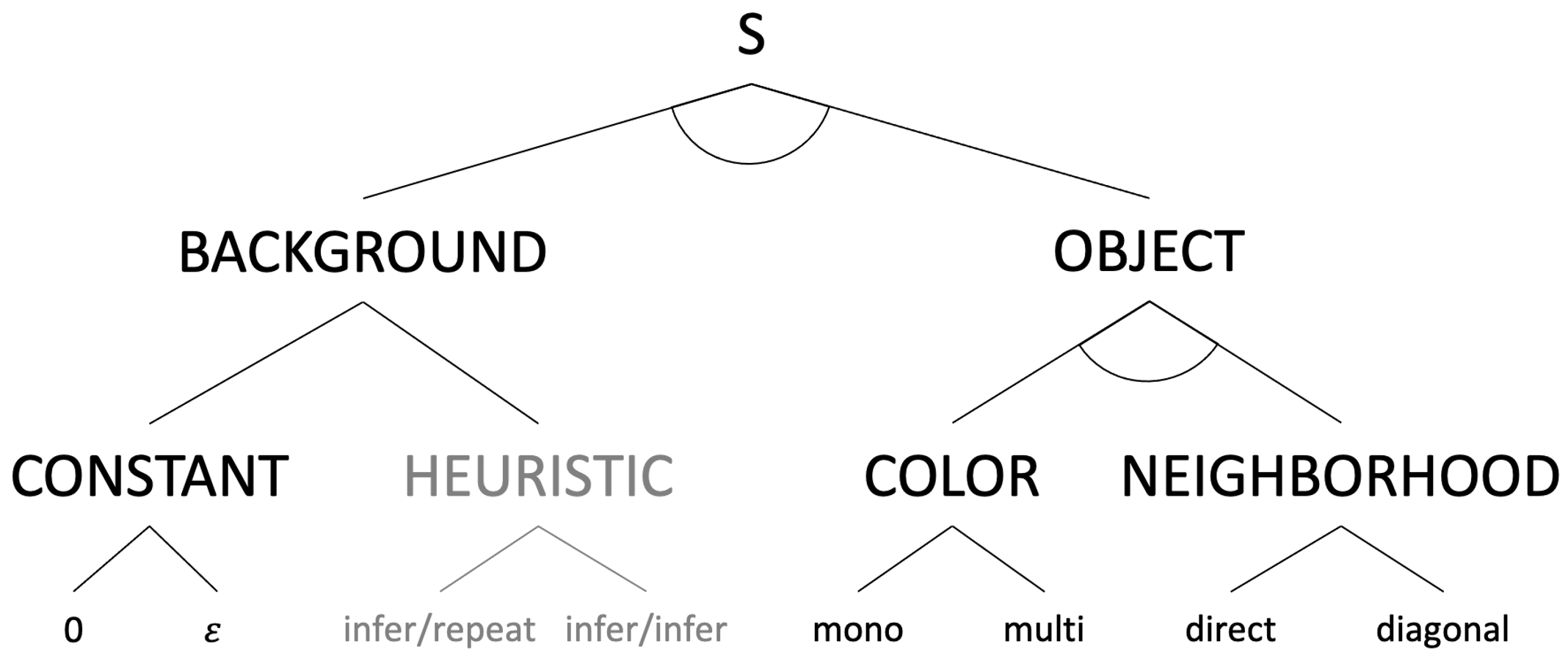}
        \caption{AND/OR graph of $\mathcal{G}_{decomp}$.}
        \label{fig:segmentationgrammar}
     \end{subfigure}
     \hfill
     \begin{subfigure}{0.48\textwidth}
         \centering
         \includegraphics[scale=0.35]{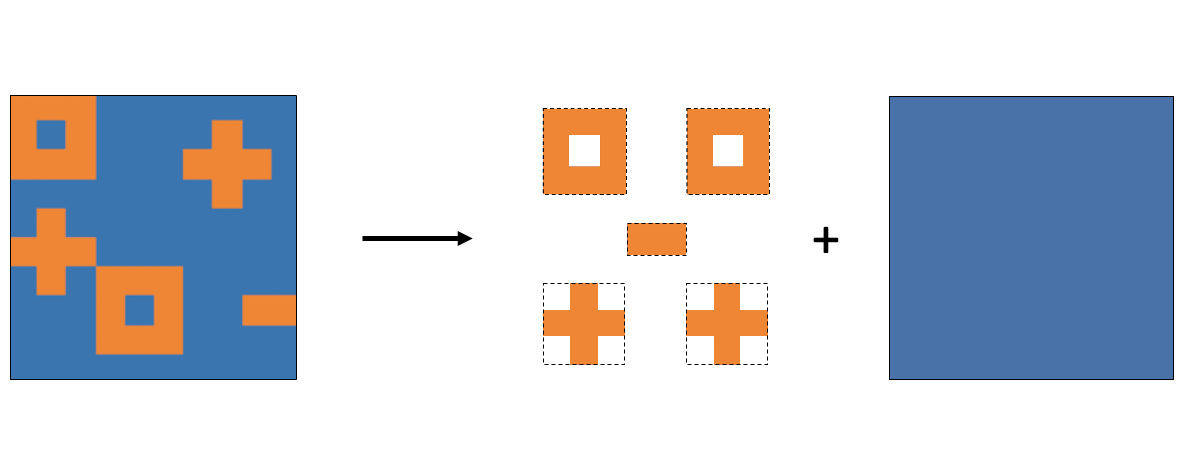}
         \caption{Application of a decomposition function $\delta$.}
         \label{fig:segmenter-0monodirect}
     \end{subfigure}
        \caption{Segmentation grammar $\mathcal{G}_{decomp}$ used for abstract visual reasoning tasks in ARC.}
\end{figure}

Our agent, BEN, works with a context-free grammar $\mathcal{G}_{decomp}= (\Psi_N, \Psi_T, \Psi_S, \mathcal{R})$ to encode these segmentation priors (visualized as an AND/OR graph in \Cref{fig:segmentationgrammar} for the domain of abstract visual reasoning). $\Psi_N$ is the set of non-terminals (e.g. background, object), $\Psi_T$ is the set of terminals (constraints), $\Psi_S$ is the start symbol and $\mathcal{R}$ the set of production rules to derive a unique set of constraints.
The language of ${\mathcal{G}_{decomp}}$ is the set of decomposition functions $\{\delta \in \Psi_T^* | \Psi_S \Rightarrow^*_{\mathcal{G}_{decomp}} \delta\}$ derivable from $\mathcal{G}_{decomp}$ under the transitive closure $\Rightarrow^*_{\mathcal{G}_{decomp}}$.
For example, the expression $\mathcal{D} \llbracket c_1,c_2 \rrbracket$ with constraints $c_1 \triangleq$ 'mono-colored' and $c_2 \triangleq$ 'direct neighbors' is an abstraction over decomposition functions that produce components containing only equally colored, directly neighboring pixels.
A decomposition function $\delta$ is applied to a bitmap by evaluating its constraints $\{c_i\}_{i=0}^{Z}$ on pairs of pixels, \Cref{equation:segmentation}. Any two pixels that satisfy all constraints are merged into the same object. The subsequent synthesis is performed on these objects instead of individual pixels.

\begin{equation}
\label{equation:segmentation}
\bigwedge_{i=0}^{Z}{\llbracket c_i \rrbracket(pixel_1,pixel_2)} = \top \rightarrow (pixel_1,pixel_2) \in object
\end{equation}

In the running task, the decomposition function learned by BEN segments the first example into four orange objects and a blue background (\Cref{fig:segmenter-0monodirect}). In order to deal with natural background/foreground separation, our domain grammar ${\mathcal{G}_{decomp}}$ treats the background object as a dense square which is partly occluded by objects in the foreground compared to a sparse square with holes ('Law of Prägnanz' from Gestalt psychology \cite{wagemans_century_2012}) (\Cref{fig:segmentationgrammar}). Its color is either constant or inferred using simple heuristics which exploit the fact that backgrounds often make up most of an image.
Abstract visual reasoning tasks frequently make use of occlusion to create the illusion of a depth ordering amongst objects. Accepting occlusion as one fundamental principle of our world model, leads to simpler object parses and later on also simpler solution programs.

In order to work with other domains, we update $\mathcal{G}_{decomp}$ to reflect the compositionality priors of the new domain. For instance, string manipulation tasks have intuitive segmentation boundaries in the form of special characters (e.g. comma, whitespace, colon). The domain of string manipulation tasks is introduced in the experiments section.

\subsubsection{Optimization - generalization difficulty}
We recall that the goal of the \textit{divide} stage is to identify components which lead the program search to discover a set of transformation programs $\mathcal{T}$ that are compact and successfully reconstruct the task outputs. A naive approach could use the domain grammar as described in \Cref{sec:segmentation} to linearly search through $\mathcal{L}_{\mathcal{G}_{decomp}}$ until it finds a segmentation which yields a perfect reconstruction of the output given some library of transformations. However, the number of decomposition functions in $\mathcal{L}_{\mathcal{G}_{decomp}}$ is too large to be traversed exhaustively. Even the evaluation of a single $\delta$ is costly because it requires us to minimize the negative log-likelihood of $P(Q,\delta,\mathcal{T})$ by doing best-first search over a potentially large number of pairwise correspondences.

The problem is further complicated by the fact that only slight syntactical differences in the segmentation constraints can lead to segmentations with vastly different semantics (usefulness in terms of the downstream synthesis). This means that the evaluation feedback from one $\delta$ cannot easily be used to direct the search towards more promising decomposition functions.
Instead, we leverage the observation that decomposition functions that produce meaningful segmentations and lead to perfect reconstruction on the first input/output pair $(I_1,R_1)$, are more likely to also yield meaningful segmentations on the remaining pairs $q_l$. The same argument applies to individual components: decomposition functions which lead to the successful reconstruction of the first component in the first output are more likely to also yield meaningful segmentations for the remaining components and examples.
This means, we estimate the usefulness of a decomposition function $\delta$ by evaluating its joint probability on a single segmented component in the first output. We apply this estimation to all decomposition functions within $\mathcal{G}_{decomp}$.

\raggedbottom

\begin{equation}
\label{equation:segmentationopt}
\hat{P}(Q,\delta,\mathcal{T}) = P((I_1,o_1),\delta,\mathcal{T}) \quad and \quad o_1 \in R_1
\end{equation}

\subsection{\textit{Align} - Structural Alignment}
\label{sec:align}

The synthesis goal is to learn a minimal set of transformation programs $\mathcal{T}$ which act on input components and reconstruct output components. We denote this partial mapping as a set of learned correspondences  $corr: o_i \rightarrow o_j$ in the Cartesian product $\delta(I_l) \bigtimes \delta(R_l)$.
This step is combinatorially expensive in theory. However, compositional real world domains provide rich constraints that we use to find likely correspondences fast.

Our approach is inspired by analogical reasoning which has been investigated as a foundational mechanism humans use to map knowledge of familiar situations onto new domains. The field's most prevalent theory is structure-mapping theory (SMT) \cite{gentner_structure-mapping_1983}, first developed in the cognitive sciences and afterwards implemented as a computational model in the computer sciences, called the structure mapping engine (SME) \cite{falkenhainer_structure-mapping_1989}. We leverage SME to search for a structural alignment between input/output examples which maximizes their shared relational structure. From there, we rank pairwise correspondences between components in the input/output relative to their contribution to the structural alignment and explore them iteratively until we have recovered enough transformation programs to solve the entire output.

In the experiments, we evaluate how maximizing the joint relational structure between examples speeds up synthesis and biases the search towards well-generalizing programs.
The structural alignment performed by SME is symbolic and, thus, requires a propositional encoding of the decomposed scenes.

\subsubsection{Propositional Encoding}

\begin{figure}[t]
    \centering
    \includegraphics[width=0.95\linewidth]{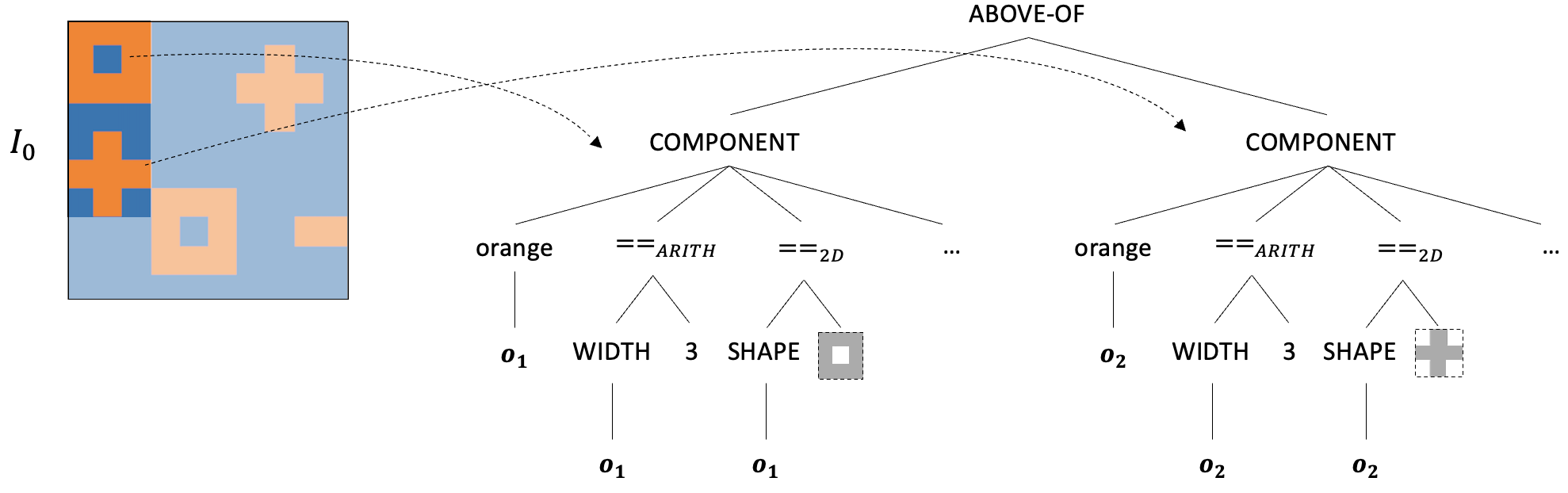}
    \caption{Example of a qualitative spatial propositional encoding of two decomposed objects in an abstract visual reasoning task.}
    \label{fig:encoding}
\end{figure}

We build on a set of basic assumptions that make the symbolic encoding of scenes domain-independent.
A propositional encoding is a directed acyclic graph (DAG) which relates expressions about components within a scene. Expressions are either primitive entities (components) or predicates. Every predicate is either a relation (i.e \textit{above-of}), a function (e.g. \textit{width}) or an attribute (e.g. \textit{orange}).
An expression $E$ is a vertex in a graph and forms edges to all of its arguments (\Cref{fig:encoding}). These are its descendants. They share $E$ as a common ancestor. An expression with no ancestors is called a root. A propositional scene encoding can contain more than one root. An expression $E'$ is reachable from an expression $E$ if it is part of its transitive closure $\mathcal{R^*(E)}$. The depth of an expression is simply the minimum number of edges needed to reach it, starting at a root node.

The objective of the encoding is to facilitate a fast alignment of the shared structure in the input/output examples which in turn leads to simpler solution programs. This is because the more similar a pair of aligned components across input/output, the shorter the program that is needed to transform one into the other.
Structure-mapping theory (SMT) argues that humans solve the combinatorial explosion of possible mappings between a base and target through \textit{structural} alignment.
This means they favor mappings between the base and target with deep shared relational structure (systematicity principle). Evidence from the cognitive sciences suggests that especially qualitative spatial relations are crucial to this process \cite{lovett_modeling_2017}.

The specific relations used are domain-dependent, e.g. basic positional (e.g. \textit{left-of}) and topological relations from the Region Connection Calculus (RCC8).
The complexity of a program that transforms one component into another is influenced both by the degree of matched relational structure and the extent of feature overlap between the two components. Therefore, we also add object features to the encoding.

\begin{table}[ht]
\centering
\footnotesize
  \begin{tabular}{c l l l}
  \toprule
  \multicolumn{4}{c}{}\\
	& \multicolumn{2}{l}{\textbf{Object features - Abstract visual reasoning}} & \\
	\cline{2-4}
	\multicolumn{4}{c}{}\\
    ID & Feature & Explanation & Derived from \\ \hline
    \multicolumn{4}{c}{}\\
    1 & color & & \\
    2 & num\_colors & number of unique colors in the object & color \\
    3 & row\_origin\_bbox & top left corner of the object's bounding box & \\
    4 & col\_origin\_bbox & top left corner of the object's bounding box & \\
    5 & shape & 2D array of pixels in bounding box & \\
    6 & size & number of pixels in the object & shape \\
    7 & width & & shape \\
    8 & height & & shape \\
    9 & ranked\_color & ordinal: decreasing with the number of objects of this color & color \\
    10 & ranked\_color\_rev & ordinal: increasing with the number of objects of this color & color \\
    11 & ranked\_size & ordinal: decreasing with size & size \\
    12 & ranked\_size\_rev & ordinal: increasing with size & size \\
    13 & ranked\_shape & ordinal: decreasing with the number of objects of this shape & shape \\
    14 & ranked\_shape\_rev & ordinal: increasing with the number of objects of this shape & shape \\
    15 & filled & boolean flag: bbox is filled by the object completely & shape \\
    \bottomrule
    \multicolumn{4}{c}{}\\
  \end{tabular}
  \captionsetup{justification=centering}
  \caption{Object features used in BEN on abstract visual reasoning tasks.}
  \label{tab:features_ARC}
 \end{table}

Nominal domain features (e.g. a component's color) are encoded as attributes. Any other features, such as a component's width or shape, are encoded as functions. For each functional output type (e.g. scalar, matrix), there is an equivalence relation (e.g. $=_{ARITH}$, $=_{2D}$) that compares specific values with constants of that type (\Cref{fig:encoding}).
Domain predicates are manually provided. We use 15 features (\Cref{tab:features_ARC}) for the abstract visual reasoning data sets.

\subsubsection{Structure-Mapping}

We provide the propositional encodings to SME and follow its three-step evaluation described in the background to extract correspondences between input/output components.
In the running example (\Cref{fig:architecture}), a local match hypothesis $mh$ (\Cref{par:structuremapping}) is formed between the \squareshape{}-shaped object in the input and the green \crossshape{}-shaped object in the output. During the second step, we derive maximal and structurally consistent global mappings between the input/output (isomorphic subgraphs in their DAGs) (line 1 in \Cref{alg:SME}).
Finally, we adapt how SME scores GMAPs to take into account both matched relational structure as well as feature similarity between matched components (line 5 in \Cref{alg:SME}). We do this to guide the downstream synthesis towards compact transformation programs. The updated scoring function \Cref{eq:SMEscore}, therefore, evaluates local match hypotheses based on their depth $d$ in the DAG and a feature similarity metric $sim$ where more importance is given to the matched relational structure $\omega_0 > \omega_1$. The score of a GMAP is equal to the sum of the scores of its match hypotheses. 

\begin{equation}
    \label{eq:SMEscore}
    score(GMAP) = \sum_{f=0}^{|root \in GMAP|}\sum_{g=0}^{|mh \in \mathcal{R}^{*}(root_f)|}{
    \omega_0*d(mh_g) + \omega_1*sim(mh_g)
    }
    \end{equation}

\begin{algorithm}[ht]
\caption{RankCorrespondences}
\label{alg:SME}
\textbf{Input:} base representation $\mathcal{E_I}$, target representation $\mathcal{E_R}$\\
\textbf{Parameters:} match constructor rules $MHC$\\
\textbf{Output:} ranked correspondences $CORR$
\begin{algorithmic}[1]

\State $GMAP = SME(\mathcal{E_I}, \mathcal{E_R}, MHC)$

\State $CORR \leftarrow \{\}$
\For {$o_j \in \mathcal{E_R}$}
\State $CORR[o_j] \leftarrow \{\}$

\For {$gmap\in$ GMAP}
\State $score = Score(gmap)$
\Comment{see \Cref{eq:SMEscore}}
\State $CORR[o_j] \leftarrow \{ CORR[o_j] \cup (o_i, score) | (o_i,o_j) \in gmap \}$
\EndFor

\State Sort $CORR[o_j]$ in decreasing order of $score$
\State Backfill $CORR[o_j]$ with missing correspondences from $\{(o_i,o_j)|o_i\in \mathcal{E_I}\}$

\EndFor

\State \Return $CORR$

\end{algorithmic}
\end{algorithm}

In the running example in \Cref{fig:exampletask_ARC}, input objects are preferably matched with their position-invariant counterparts in the output.
These pairwise correspondences receive high evaluation scores because they map most of the qualitative relational structure in the input to the output. Correspondences between, e.g. the orange \crossshape{}-shaped objects in the input and green \crossshape{}-shaped objects in the output, are also generated due to feature similarities. However, their scores are much lower as they do not account for any of the relational structure shared between the input/output.

In the following \textit{conquer} stage, we perform synthesis on individual pairs of component correspondences extracted from GMAPs. Those correspondences are explored in the order of their associated GMAP scores (line 6 in \Cref{alg:SME}). This means that we greedily perform synthesis on input/output pairs with similar relational structure and features. However, the best-first search will never cause DA\&C to miss a transformation program because $CORR$ contains an exhaustive list of all pairwise correspondences of an output object with the input (line 10 in \Cref{alg:SME}). In the worst case, it will iteratively explore the Cartesian product of component correspondences between the input/output.

Line 1 in \Cref{alg:SME} calls SME (Structure Mapping Engine). We refrain from reprinting detailed pseudo code in this paper and refer to \cite{falkenhainer_structure-mapping_1989} for the original implementation of SME.

\paragraph{Complexity analysis.} (1) The construction of local match hypotheses has a worst-case performance of $\mathcal{O}(N^2)$ with $N$ being the average number of expressions in $\mathcal{E}(I)$ and $\mathcal{E}(R)$. Match constructor rules are applied to every predicate pair in the base/target. (2) The worst-case performance of finding maximal structurally consistent GMAPs is that of finding maximal sets of isomorphic subgraphs between the base/target which is given by $\mathcal{O}(N!)$. (3) The computation of a GMAP evaluation score, \Cref{eq:SMEscore}, is a graph walk over the number of roots within each GMAP and each of their transitive closures $\mathcal{R^*}(root)$. Its performance is bounded by the total number of local match hypotheses which is $\mathcal{O}(N^2)$ in the worst case. The initial generation of the encodings (using unary and binary predicates) has a worst-case performance of $\mathcal{O}(N^2)$ which does not change the worst-case performance of the entire algorithm.
In practice, we observe that the structural alignment using SME (\Cref{alg:SME}) performs significantly below its worst-case bound and depends heavily on the propositional scene encoding. Structural alignment performs best on deeply nested relational encodings with a variety of different relations \cite{falkenhainer_structure-mapping_1989}. We analyze the impact of analogical reasoning as part of the ablation studies in the experiments section.

\subsection{\textit{Conquer} - Synthesis Of Transformation Programs}

The structural alignment of two scenes yields a ranked list of component tuples $(o_i,o_j)$ between the input and output scenes.
From this, the search for transformation rules $\tau$ proceeds in two steps. (1) Rule consequent: We treat each tuple as its own synthesis specification where the goal is to learn a transformation program $p'(o_i) = o_j$ that transforms $o_i$ into $o_j$. (2) Rule condition: For every transformation program $p'$, we learn a concept $C_{p'}$ over all input components that make use of $p'$ across examples. A transformation rule, therefore, consists of a specific manipulation and the context in which it is applied.

BEN repeatedly picks the highest-ranking component tuple (as evaluated in the \textit{align} stage), solves a synthesis task after which it updates the information on (1) known transformations and (2) their respective concepts. At the end of each iteration, if the set of collected transformations up to this point is sufficient to reconstruct all components across all outputs, it provides a solution program that consists of a minimal set of transformations. It can update its solution program as more compact transformation rules are discovered until it eventually terminates once synthesis has been performed on all correspondences in the Cartesian product $\delta(I_l) \bigtimes \delta(R_l)$.

We now consider a single synthesis loop (line 12 in \Cref{alg:divide-align-conquer_overview}) on the highest-ranking correspondence (component tuple).

\begin{figure}
    \footnotesize
    \caption{Domain grammar for abstract visual reasoning $\mathcal{G}_{ARC}$.}
    \label{fig:Gdomain_ARC}

    \begin{tabular}{ | >{\raggedleft}p{0.2\textwidth} p{0.05\textwidth} p{0.665\textwidth}| }
        \hline
        $\tau$ & $\rightarrow$ & (if \textcolor{Blue}{CONCEPT} then \textcolor{Red}{TRANSFORM})\\
        \multicolumn{3}{|l|}{}\\
        TRANSFORM & $\rightarrow$ & \makecell[tl]{oi, border(TRANSFORM, corr, env, \#hole),\\inner(TRANSFORM, corr, env),\\color(TRANSFORM, corr, env, oj.color),\\shape(TRANSFORM, corr, env, oj.bbox),\\replace(TRANSFORM, corr, env, select(REFERENCE)),\\cut(TRANSFORM, corr, env, \#hole),\\denoise(TRANSFORM, corr, env),\\move(TRASNFORM, corr, env, MD, \#hole),\\scale(TRANSFORM, corr, env, \#hole),\\rotate(TRANSFORM, corr, env, \#hole),\\mirror(TRANSFORM, corr, env, \#hole),\\complement(TRANSFORM, corr, env)}\\
        \multicolumn{3}{|l|}{}\\
        MD & $\rightarrow$ & by, to, in\\[0.05cm]
        REFERENCE & $\rightarrow$ & most\_colorful, largest, other\\
        \hline
    \end{tabular}
\end{figure}

\subsubsection{Rule Consequent: Search For A Transformation}

The \textit{conquer} stage can use an off-the-shelf synthesis engine.
BEN for example follows a generate \& test approach using top-down enumerative search that explores all programs up to a predefined depth $d$.
Transformation programs in the running example are derived from the domain grammar $\mathcal{G}_{ARC}$ in \Cref{fig:Gdomain_ARC} which contains primitives for basic geometric operations such as scaling, translating, rotating, and filling of objects.

\paragraph{Optimizations.}
Here we introduce a crucial adaptation to standard enumerative search in order to exploit the fact that synthesis is executed on component tuples: The domain grammar $\mathcal{G}_{ARC}$ which houses the geometric primitives used to reason about abstract visual scenes does not contain derivations for the arguments of its primitives. Their arguments are either analytically specified by directly referencing the information contained within the output component of the underlying correspondence (e.g. see the 'oj.color' in the color() primitive) or they make use of 'holes' which are dynamically filled once a program candidate is executed. When a primitive is called with a 'hole' as argument, it derives a parametrization which is correct in the context of the current correspondence or returns an error upon which the program candidate is discarded immediately. For example, the scale primitive analytically computes the scaling factor that is needed in order to change the size of the input component to that of the output component and replaces the 'hole' by this value. The use of 'holes' significantly reduces the size of the search space because fewer programs are enumerated. In the experiments, we show that by exploiting the information that is available within the component tuple in this way, the speed-up in the synthesis can well account for the overhead needed to identify and encode meaningful component tuples.

Depending on the type of arguments and their permissible range of values, 'holes' allow us to work with primitives such as the 'shape' operator which would otherwise be intractable to search through: Consider the transformation program \textit{o.shape(\crossshape{}).color(green)} which is the result of the synthesis performed on the \squareshape{}-shaped component in the input and the green \crossshape{}-shaped component in the output. The 2D shape argument can simply be deduced from the matched output component and does not need to be searched. The same is true for the color argument.

Performing synthesis on a single component tuple potentially leads to a large number of successful transformation programs many of which do not generalize to other examples. In order to bias the synthesis towards minimal and well-generalizing solution programs, after each synthesis loop, we only keep track of whatever transformation program reconstructed the most additional output components across examples. This is the reason why in line 12 in \Cref{alg:divide-align-conquer_overview} the synthesis only returns a single transformation program. The dictionary $\mathcal{T'}$ initialized in line 5 in \Cref{alg:divide-align-conquer_overview} records and updates which components are solved by which transformation program.

\subsubsection{Rule Condition: Learn A Concept Definition}
\label{sec:conceptlearning}

In the previous step, we synthesized transformation programs on input/output component tuples. In order to combine these individual transformations into a program that solves the entire task, we now learn Boolean logic formula that describe the context in which a transformation is to be executed. Afterwards, we will use these to combine transformation programs with 'if-then' statements.

The problem is a standard binary concept learning task over components $f(o_i) \rightarrow \{0,1\}$ where the concept to be learned represents a subset of components.
For each transformation program $p'$, we partition the set of input components into three groups: 
\begin{enumerate}
    \item The set of positive components $P$ that successfully reconstruct a component in the output using $p'$, $\forall o_i \in P: p'(o_i) \in \{\delta(R_l)\}_{l=0}^{L}$.
    \item The set of negative components $N$ that incorrectly reconstruct parts of the output if $p'$ was applied to a component from this set.
    \item A set of neutral objects which do not generate false outputs but also don't reconstruct any component in the output if $p'$ was applied to a component from this set (e.g. partial reconstruction of an output component, out-of-bounds transformations).
\end{enumerate}

\begin{algorithm}
\caption{Constrained DNF learner}
\label{alg:cDNF}

\textbf{Input:} $X$: component representations $\{0,1\}^{(n,m)}$, $Y$: labels of $\{0,1\}^n$ \\
\textbf{Parameters:} $j$: max number of conjunctions \\
\textbf{Output:} $DNF$
    
    \begin{algorithmic}[1]
    \State{$DNF = []$}
    \State{$P^+ = \{o_i|o_i\in X \land Y[o_i] == 1\}$}
    \State{$N^- = \{o_i|o_i\in X \land Y[o_i] == 0\}$}
    \For{i in 1..$j$}
    	\State{$conj$ $\leftarrow$ solve Equation 6-9 on $P^+$ and $N^-$}
    	\State{Add $conj$ to $DNF$}
    	\State{Remove from $P^+$ all components covered by $conj$}
    	\If{$P^+$ is empty}
    	    \State{\Return{$DNF$}}
    	    \Comment{Perfect classifier, done}
    	\EndIf{}
    \EndFor
    \State{\Return{$DNF$}}
    \end{algorithmic}
\end{algorithm}

A key challenge is the limited number of observations available per task. There are only as many examples as there are components in all inputs, generally between 3 to 30. However, the fact that a task requires perfect reconstruction of all outputs introduces a strong inductive bias: Namely, the concept must cover all positive components $P$ but none of the negative components $Z$.
If the concept did include even a single negative example, at least one of the example outputs would be reconstructed incorrectly. The learner imposes no constraints on neutral components.
In addition, we expect a good concept to only use a small number of component features for generalization purposes.

Given that we have already generated symbolic encodings of each component during the \textit{align} phase, we learn a Disjunctive Normal Form (DNF) \cite{valiant_learning_1985} over component features. A DNF is a disjunction of conjunctions. It is best understood as a set of rules, where each rule specifies a set of properties. If at least one rule applies, the DNF evaluates to true. It is a function $\{0,1\}^m \rightarrow \{0,1\}$.
In order to learn a DNF over a variety of domain features, we first hash any non-numeric features (e.g. color, shape) and then double one-hot encode all attributes, such that there is a Boolean for each attribute-value combination as well as its negation (e.g. 'color == orange' and 'color != orange'; \mbox{'shape == \crossshape'} and \mbox{'shape != \crossshape'} etc).

Instead of enumerating all conjunctions up to a fixed number of conjuncts, and selecting from those, we use an \textit{implicit} generation approach where we formulate the problem of generating a single conjunction through constraint optimization. We make use of constraint programming for item set mining \cite{guns_k-pattern_2013} to formulate the following constrained optimization problem:

\begin{align} 
 maximize_{S} \quad &w\sum O^+ - \sum S \label{DNF:objectivefunction} \\
     s.t. \quad &O^+ = cover(S, P) \label{DNF:Pcovered} \\ 
     	   &O^- = cover(S, N) \label{DNF:Ncovered} \\
     	   &\sum O^- = 0 \label{DNF:constraint}
\end{align}

where $S$ ($|S|=m$) contains a Boolean decision variable for every Boolean attribute in the double one-hot encoded component representation, $P$ and $N$ are the positive and negative components (their representations), $O^+$ and $O^-$ contain a Boolean decision variable for every positive/negative component that represents whether the component is covered by the conjunction $S$ or not. 
The constraints on lines \eqref{DNF:Pcovered} and \eqref{DNF:Ncovered} compute which objects in $P$ and $N$ are covered by $S$. The constraint on line \eqref{DNF:constraint} ensures no negative components are covered and the objective function on line \eqref{DNF:objectivefunction} is a lexicographic optimization which first maximizes the number of covered $P$ and then minimizes the number of Boolean attributes used in the conjunction. The pseudocode of the overall DNF learner is in \Cref{alg:cDNF}.

While the optimization problem has to search over a worst-case exponential number of conjunctions, there are only a few examples and the constraints provided by negative instances greatly limit the search space. In practice, constraint solvers find solutions to this problem very rapidly. In the running example (\Cref{fig:exampletask_ARC}), the learned DNFs contain a single conjunction with a single Boolean attribute set to true.

\begin{Verbatim}[fontsize=\small, commandchars=\\\{\}]
if o.shape == \squareshape then o.replace_by(\crossshape).color(green)
if o.shape != \squareshape then o.identity()
\end{Verbatim}

To summarize, a transformation rule consists of a condition and consequent. Conditions are concepts about sets of components. Consequents are transformation programs over domain-specific transformation primitives. Multiple transformation rules together with a decomposition function make up a solution program. When a solution program is executed on a test input, the input is first decomposed into components which are then evaluated against each of the transformation rules of the program. Whenever a rule condition evaluates to true, it triggers the execution of the corresponding transformation program on the current input component which completes the DA\&C paradigm.

\section{Experiments}
\label{sec:experiments}

Following our description of the DA\&C paradigm, we claim that the use of segmentation and analogical matching for structured domains enables agents to learn complex programs in less time. To this end, we evaluate BEN, our implementation of DA\&C using standard top-down enumerative synthesis, across two domains: abstract visual reasoning used as a running example throughout the paper (\Cref{sec:dom2}) and string transformation tasks (\Cref{sec:dom1}). We select those because they cover the spectrum from an established experimentation domain for program synthesis (string transformations) to a challenging new domain benchmark that has proven difficult for current synthesis strategies. Our experiments seek to investigate three research questions:

\begin{itemize}
    \item[\textit{\textbf{Q1}}] How does the predictive accuracy of BEN behave in comparison to state-of-the-art ILP systems? We pay special attention to how their predictive accuracies are influenced by the size of the example sets.
    \item[\textit{\textbf{Q2}}] How does the use of analogical matching on structured real-world synthesis tasks influence the runtime complexity of BEN which is expected to run with a worst case time complexity of $O(n^2)$ with $n$ being the number of segmented components in the example input/output?
    \item[\textit{\textbf{Q3}}] What is the effect of growing solution programs on solution times of BEN, especially in the context of real-world structured domains in which solution programs are frequently observed to linearly grow in the number of independent sub programs?
\end{itemize}

All experiments are run on a desktop with a single M1 Max CPU and 64GB of RAM.
Our code and an overview of the tasks solved by BEN are available in the supplementary materials.

\begin{table*}[!b]
\centering
\footnotesize
  \begin{tabular}{c l l l}
  \toprule
  \multicolumn{4}{c}{}\\
	& \multicolumn{3}{l}{\textbf{Object features - String transformation tasks}} \\
	\cline{2-4}
	\multicolumn{4}{c}{}\\
    ID & Feature & Explanation & Derived from \\ \hline
    \multicolumn{4}{c}{}\\
    1 & content & list of characters & \\
    2 & index\_front & index position of substring from the front & \\
    3 & index\_back & index position of substring from the back & \\
    4 & index\_even & boolean: even front index & index\_front \\
    5 & length & number of characters & content \\
    6 & number\_of\_uppers & number of capitalized characters & content \\
    7 & number\_of\_lowers & number of lowercase characters & content \\
    8 & number\_of\_digits & number of digits & content \\
    9 & number\_of\_alphas & number of alphabetical characters & content \\
    10 & number\_of\_alnums & number of alphanumeric characters & content \\
    11 & all\_upper & boolean: all capitalized characters & content \\
    12 & all\_lower & boolean: all lowercase characters & content \\
    13 & all\_digits & boolean: all digits & content \\
    14 & starts\_with\_upper & starts with capitalized character & content \\
    \bottomrule
    \multicolumn{4}{c}{}\\
  \end{tabular}
  \captionsetup{justification=centering}
  \caption{Object features used in BEN on string transformation tasks.}
  \label{tab:features_strings}
\end{table*}

\subsection{Experiment 1: String Transformations}
\label{sec:dom1}

We evaluate our approach using the established domain of string transformation tasks.

\paragraph{Materials.}
We experiment with a publicly available data set of 130 real world string transformation tasks from \citeA{cropper_learning_2020}. The initial subset of tasks was curated by \citeA{gulwani_automating_2011} from online Microsoft Excel forums, later expanded by \citeA{lin_bias_2014} with additional handcrafted spreadsheet manipulations and has been repeatedly used as a benchmark in program synthesis.

\begin{table*}[!b]
\centering
\footnotesize
  \begin{tabular}{c l l l}
  \toprule
  \multicolumn{4}{c}{}\\
	& \multicolumn{3}{l}{\textbf{Transformation primitives - String transformation tasks}} \\
	\cline{2-4}
	\multicolumn{4}{c}{}\\
    ID & Transformation & Explanation & Arguments \\ \hline
    \multicolumn{4}{c}{}\\
    1 & drop\_first(n) & drops leading n characters & n $\in \{1 ... len(o_i)\}$\\
    2 & drop\_last(n) & drops last n characters & n $\in \{1 ... len(o_i)\}$\\
    3 & take\_from\_front(n) & selects leading n characters & n $\in \{1 ... len(o_i)\}$\\
    4 & take\_from\_front(n) & selects last n characters &  n $\in \{1 ... len(o_i)\}$\\
    5 & to\_uppercase() & capitalizes string & \\
    6 & to\_lowercase() & lowercase string & \\
    7 & capitalize\_first() & capitalizes first character & \\
    8 & add\_space() & appends white space & \\
    9 & add\_dot() & appends dot & \\
    10 & add\_comma() & appends comma & \\
    11 & replace(s) & replace with string s & s is given by the output string $o_j$ \\
    \bottomrule
    \multicolumn{4}{c}{}\\
  \end{tabular}
  \caption{Transformation primitives used in BEN on string transformation tasks.}
  \label{tab:transformationprimitives_strings}
 \end{table*}

For this experiment, we provide BEN access to string features (\Cref{tab:features_strings}) and primitives used to manipulate character sequences (\Cref{tab:transformationprimitives_strings}). We make sure all agents are evaluated using a set of similar manipulation primitives.
We define the decomposition grammar $\mathcal{G}_{decomp}$ over subsets of non-alphanumerical characters (e.g. whitespace, comma, dot). A decomposition function $\delta$ is one that splits a string based on the characters defined in this subset. In addition, we also consider numeric characters as possible delimiters and include a total decomposition that splits strings into individual characters. We chose this decomposition grammar because it reflects much of the compositional structure of real world strings. At the same time, it is already too expressive to be searched by naive enumeration.  
During decomposition, delimiters are appended to the preceding component. We include \textit{drop\_last()} as a primitive operation into BEN's knowledge base which can be used to access components without the trailing delimiter.

\begin{figure}[!t]
     \centering
     \begin{subfigure}{0.49\linewidth}
        \centering
        \includegraphics[width=\linewidth]{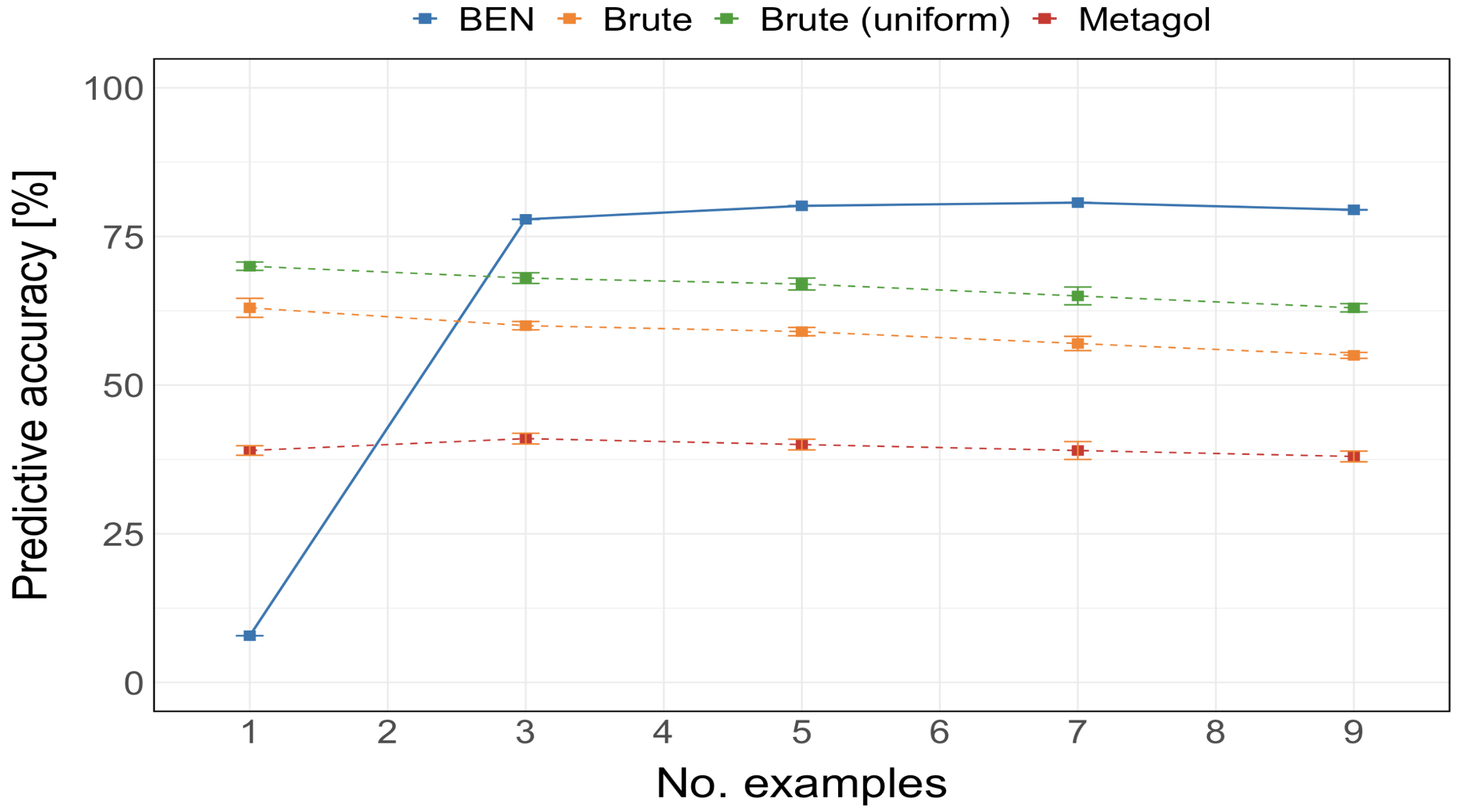}
        \caption{Predictive accuracy.}
        \label{fig:exp1_predictiveaccuracy}
     \end{subfigure}
     \hfill
     \begin{subfigure}{0.49\linewidth}
        \centering
        \includegraphics[width=\linewidth]{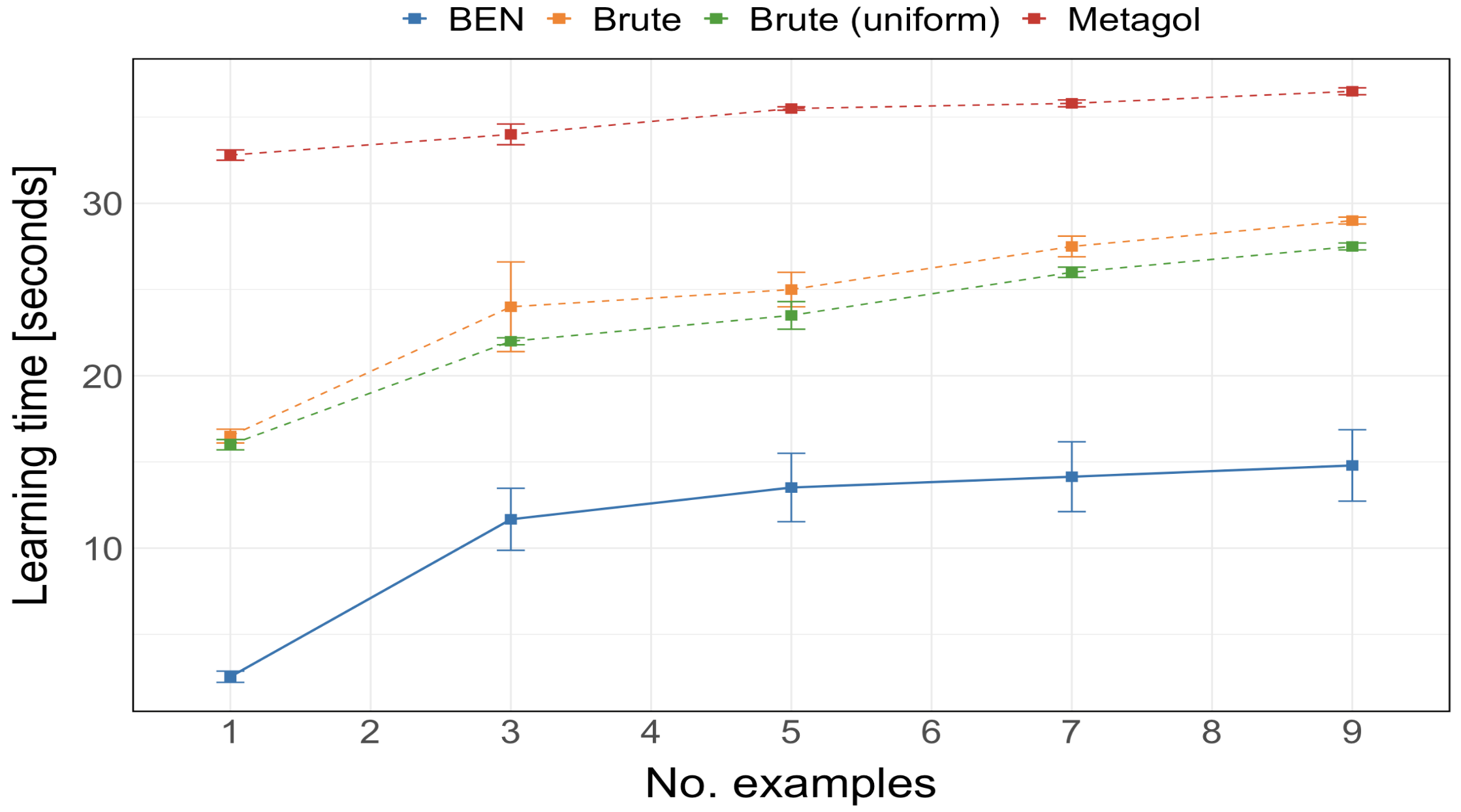}
        \caption{Mean learning time.}
        \label{fig:exp1_learningtime}
     \end{subfigure}
        \caption{Performance on the real world string transformations data set. We report first found solutions for all agents. All agents were executed with a time budget of 60s per task.}
        \label{fig:exp1}
\end{figure}

BEN is evaluated against Brute \cite{cropper_learning_2020} and Metagol \cite{muggleton_meta-interpretive_2015,cropper_metagol_2016}, two state-of-the-art ILP systems for learning recursive programs.
\textbf{Brute} performs best-first-search guided by an example-dependent loss function. It was specifically designed for the synthesis of large programs. In addition, we also compare against a version of Brute with a traditional entailment-based loss function.
\textbf{Metagol} works with user-specified meta-rules which serve as a declarative bias on the type of clauses that are considered as hypotheses, thus, limiting the search space. We supply Metagol with the \textit{identity}, \textit{inverse}, \textit{precon}, \textit{postcon}, \textit{chain} meta-rules as recommended for learning dyadic programs by \citeA{cropper_metagol_2016}.

BEN achieves significantly higher predictive accuracy than Brute and Metagol, even though BEN cannot learn recursive programs and cannot perform predicate invention.
Its predictive accuracy is 17 percentage points above that of Brute and double that of Metagol on tasks with 9 input examples \textbf{Q1}.

BEN solves over 80\% of test examples given 2-3 training examples per task. In comparison, Brute achieves a peak predictive accuracy of only 70\%. Its peak performance is lower because Brute has to find the entire solution program within a single synthesis search which becomes exponentially more difficult in the size of the program. For example, in order to extract the CPU usage '\textit{95\%}' from a command line output '\textit{16,079 inferences, 0.003 CPU in 0.003 seconds (5842660 Lips, 95\% CPU)}', Brute learns a logic program that consists of 9 clauses and 31 literals which recursively deletes characters from the front and back of the input string until only the CPU usage remains. BEN in comparison, learns a program that first segments the input using a whitespace delimiter and extracts from it the second to last substring from which it deletes the trailing whitespace.

DA\&C leads to a speed-up in synthesis which more than makes up for the additional time needed to compute a decomposition, find a meaningful structural alignment between substrings in the input/output, and learn concepts for each transformation program. This shows in the average learning times in \Cref{fig:exp1_learningtime}. BEN consistently performs below the average learning times of both Brute and Metagol. On tasks with 9 examples, Brute runs 28s on average, Metagol 37s while it only takes BEN 15s to process a task on average. Reported learning times for BEN include the time spent searching for a segmentation, encoding the segmented components, aligning the input/output, and learning concepts for individual transformation programs.

BEN, compared to Brute and Metagol, shows a unique trend in predictive accuracy over the number of input examples. Brute reaches its peak performance on tasks with a single training example. Its performance monotonically decreases to 63\% as more training examples are added. More training examples make it more difficult to find hypotheses that cover this growing set of examples (\Cref{fig:exp1_predictiveaccuracy}). BEN, on the other hand, does not show a degradation in predictive accuracy on larger example sets. Its predictive accuracy on tasks with 3 training examples is not significantly different from its predictive accuracy on tasks with 9 training examples \textbf{Q1}. The reason for this is that BEN only performs synthesis on component tuples and not sets of training examples where the complexity of a single synthesis loop is independent of the number of total training examples in a task.
The number of examples directly impacts the second part of the \textit{conquer} stage, the concept learning. This is because in order to learn a concept, every component is assigned one of three labels (positive, negative, neutral see \Cref{sec:conceptlearning}) which requires the execution of the transformation program on each input component. The impact shows in BEN's average learning times which follow a logarithmic increase. As the number of training examples increases by a factor of three, mean learning times increase by 50\%.
Notably, established ILP methods such as Brute and Metagol don't show a saturation in learning times, because their synthesis loop requires that each program candidate gets evaluated across all training examples (\Cref{fig:exp1_learningtime}).

\begin{figure}[!b]
     \centering
     \begin{subfigure}{0.49\linewidth}
        \centering
        \includegraphics[width=\linewidth,height=4cm]{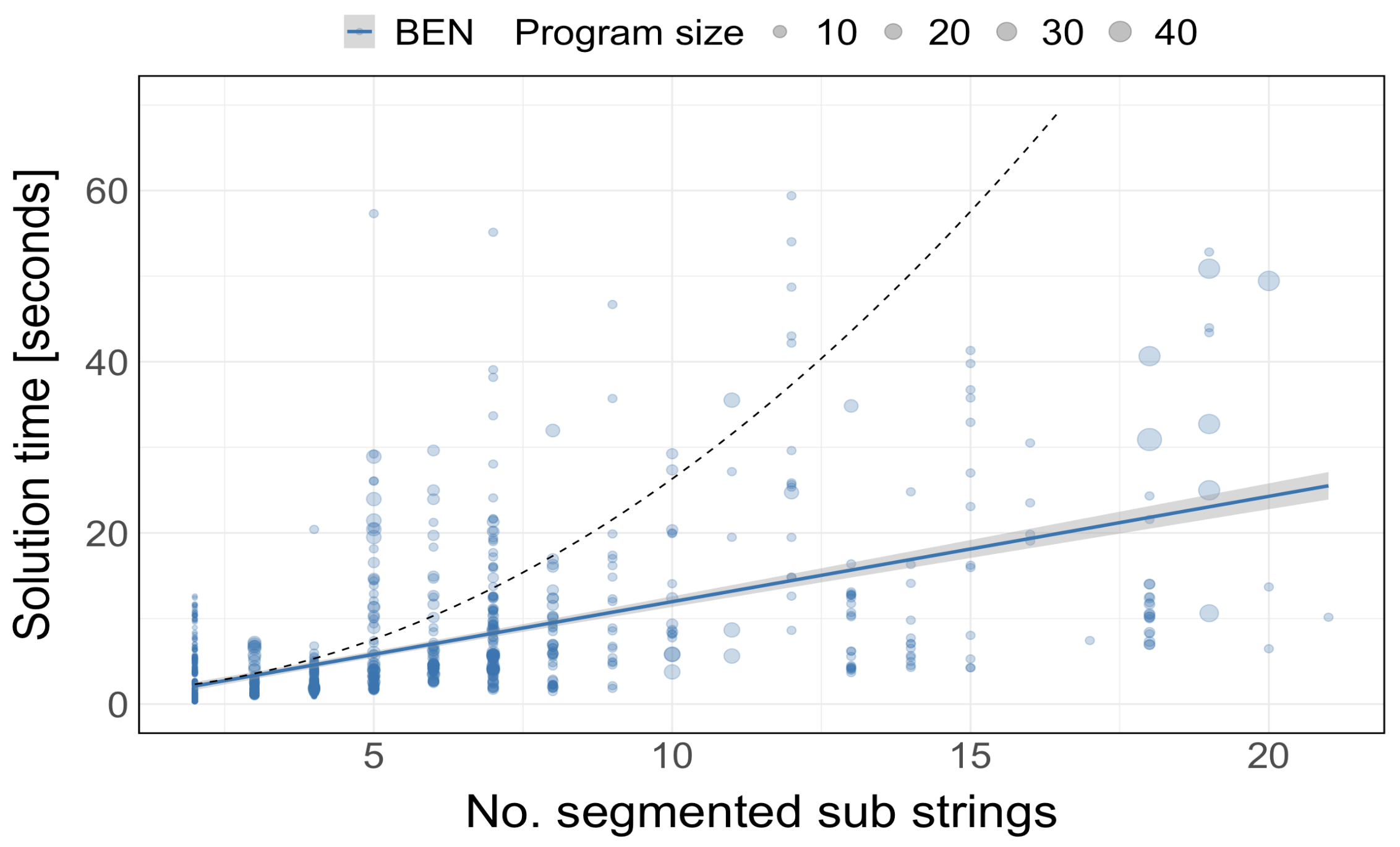}
        \caption{Solution times per segmented substrings.}
        \label{fig:exp1_Nobjects}
     \end{subfigure}
     \hfill
     \begin{subfigure}{0.49\linewidth}
        \centering
        \includegraphics[width=\linewidth,height=4cm]{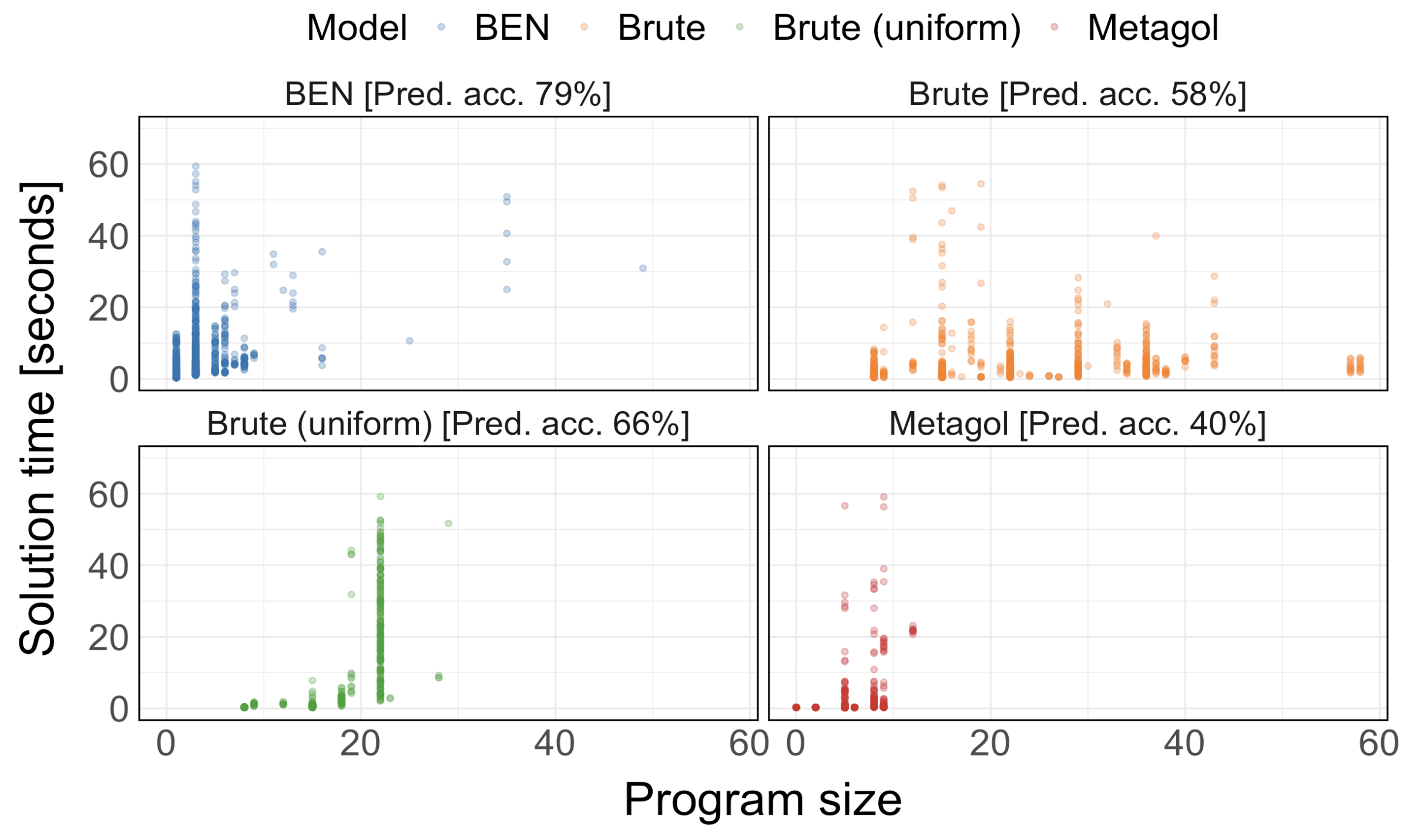}
        \caption{Solution times per program size.}
        \label{fig:exp1_programsize}
     \end{subfigure}
    \caption{Time complexity on string transformation tasks. We report first found solutions on tasks with at least 3 input examples for all agents. Only tasks solved within the time budget of 60s are considered.}
\end{figure}

If the training set consists of only a single example, BEN cannot determine how well a transformation program generalizes to other training examples (\Cref{fig:exp1_predictiveaccuracy}). As a consequence, it tends to learn overly specific transformations (e.g. extensive use of \textit{shape} operations). In much the same way, the concept learning in BEN requires a minimum amount of 2-3 positive examples in order to learn an informative selector with high generalization power to held-out test examples (and its components). Brute and Metagol perform significantly better on tasks with only a single example because they have better knowledge priors on what constitutes a generalizable program.
Missing segmentation primitives are the primary reason why BEN does not solve more tasks. BEN cannot segment examples with arbitrary regex expressions, as for instance, the letter sequence 'aabb' would be needed in order to extract the first substring from the following input 'a38bz2saabb21u17a' $\rightarrow$ 'a38bz2s'.\\
Finally, we compare empirical solution times of BEN against its theoretical worst case bound of $\mathcal{O}(n^2)$ in the number $n$ of substrings in the input and output (dashed graph in \Cref{fig:exp1_Nobjects}). The number of substrings per example significantly influences learning times even after controlling for the size of the solution program (F$(3,1137)=218.8, p<.001$) with an explained variance of 36\%. This is to be expected because examples with more components will have a larger set of plausible component tuples to search through. In practice, useful transformations are already recovered after only a few synthesis loops well below the expected theoretical worst-case bound \textbf{Q2}.
The search space of Brute (uniform with logical entailment) and Metagol exponentially grows in the size of $\mathcal{T}$, whereas the search space in BEN remains constant in size and is traversed a linear $m$ times depending on the number $m$ of transformation rules in a solution program. The total size of the solution program shows no exponential impact on learning times in BEN \textbf{Q3} (\Cref{fig:exp1_programsize}). When Brute uses an example dependent loss function instead of logical entailment, its solutions times also do not show an exponential trend with larger programs. The loss function it uses to guide the synthesis search is the reconstruction accuracy in the output string which is similar to BEN's evaluation of reconstructed substrings (components) in the output. By working with substrings instead of characters, however, BEN not only solves more tasks but also learns much shorter solution programs on average (\Cref{fig:exp1_programsize}). Even though Metagol also learns compact solutions programs fast, its predictive accuracy is only half of that of BEN which makes a direct comparison difficult.

\subsection{Experiment 2: Abstract Visual Reasoning}
\label{sec:dom2}

In order to demonstrate the effectiveness of DA\&C on program synthesis in high dimensional domains, we apply BEN to abstract visual reasoning tasks. In this setting, we compare to the state-of-the-art agents outside of ILP (\Cref{subsec:benchmark}) and evaluate different ablated baselines of BEN (\Cref{subsec:baselines}).

\paragraph{Materials.} We use the training part of the Abstraction and Reasoning Corpus (ARC) \cite{chollet_abstraction_2020} (Apache 2.0 license), which consists of 400 data sets that each contain 2-10 examples comprising an input and an output where the outputs are generated by an unspecified program that we wish to synthesize. The language of programs is determined by developers and agents themselves; the benchmark does not specify a language over programs.

\subsubsection{Performance Benchmark}
\label{subsec:benchmark}

We compare BEN against ARGA (Abstract Reasoning with Graph Abstractions) \cite{xu_graphs_2022}, an agent specifically developed for ARC. It is equipped with object centric priors which it uses to first segment examples into scenes of objects and then performs greedy best-first search over graph representations using a DSL to manipulate the 'scene graph'. We also compare against the winning submission of the ARC Kaggle competition which makes use of a custom engineered DSL with handcrafted primitives and a performance optimized implementation in C++ to perform brute-force search over bitmap manipulations \cite{wind_1st_2020}. We evaluate BEN with the object features and geometric transformation primitives introduced in the main section of this paper (\Cref{tab:features_ARC} and \Cref{fig:Gdomain_ARC}).

We first report results on the entire ARC dataset and then move on to its subsets for 'movement', 'recoloring', and 'augmentation' tasks defined by \cite{xu_graphs_2022}. The Kaggle agent solves the most tasks over all, close to 47\% within a 2 minute time budget. Its performance optimized implementation in C++ allows it to generate and test over 1.3 Mio programs per task which it does in 35s on average.
BEN solves about half as many tasks for which it only generates an average of 0.2\% as many candidate programs per task. The use of program 'holes' in the hierarchical DA\&C search provides a succinct solution space in which arguments are deduced from the components in the examples instead of being searched top-down as done by the Kaggle agent. BEN takes an average of 23s to solve a task which is competitive with the Kaggle agent despite its Python implementation. Both BEN and the ARGA agent use a significantly reduced DSL compared to the Kaggle winner. Whereas the Kaggle agent has access to 42 unique transformation primitives, many of which were handcoded to solve specific tasks, BEN and ARGA purely rely on 11 generic geometric primitives. Despite sharing the same number of primitives, BEN solves twice as many tasks and generates three times fewer candidate programs as ARGA.
BEN also consistently solves more tasks on the 'augmentation', 'movement', and 'recolor' subsets than ARGA but cannot quite reach the performance of the  Kaggle agent for the restrictive setting of a 2 min time budget per task. Notably, BEN consistently generates far fewer candidate programs than any of the other two agents. On the 'movement' and 'recolor' subsets, this even results in a reduction of the search space by one order of magnitude compared to the graph based representation of ARGA which also makes heavy use of segmentation priors.

\begin{table}[!t]
\centering
\footnotesize
\begin{tabular}{llllll}
\toprule
\multicolumn{5}{c}{}\\
Data set & Model & Number solved & Candidates explored & Avg. solution time \\
\hline \hline

\multirow{ 3}{*}{ARGA - augmentation} & \#1 Kaggle & \textbf{24/67} (35.82\%) & 1932368 & 35.90s \\
& ARGA & 15/67 (22.39\%) & 7282 & 17.51s \\
& BEN & 19/67 (28.36\%) & 6818 & 29.60s \\
\hline

\multirow{ 3}{*}{ARGA - movement} & \#1 Kaggle & \textbf{19/31} (61.29\%) & 1906875 & 34.48s \\
& ARGA & 9/31 (29.03\%) & 12233 & 14.22s \\
& BEN & 13/31 (41.94\%) & 1743 & 17.83s \\
\hline

\multirow{ 3}{*}{ARGA - recolor} & \#1 Kaggle & \textbf{23/62} (37.10\%) & 1571957 & 30.10s \\
& ARGA & 18/62 (29.03\%) & 18851 & 25.63s \\
& BEN & 22/62 (35.48\%) & 2304 & 23.42s \\
\hline

\multirow{ 3}{*}{All} & \#1 Kaggle & \textbf{186/400} (46.50\%) & 1293492 & 34.65s \\
& ARGA & 45/400 (11.50\%) & 12412 & 19.75s \\
& BEN & 90/400 (22.50\%) & 3311 & 23.20s \\
\hline

\bottomrule
\end{tabular}
\caption{Results on the training part of the Abstraction and Reasoning Corpus (ARC). First found solutions within a maximum time budget of 2 minutes per task reported. The \#1 Kaggle agent and BEN were both run at a search depth of 4 on all tasks. During the actual Kaggle competition, the \#1 Kaggle agent ran as an ensemble at different search depths to optimize scheduling. We report more challenging results at a search depth of 4. The ARGA agent does not expect a search depth as input; we use a time budget of 2 minutes.}
\end{table}

\begin{figure}[t]
     \centering
     \begin{subfigure}{0.49\linewidth}
        \centering
        \includegraphics[height=28mm]{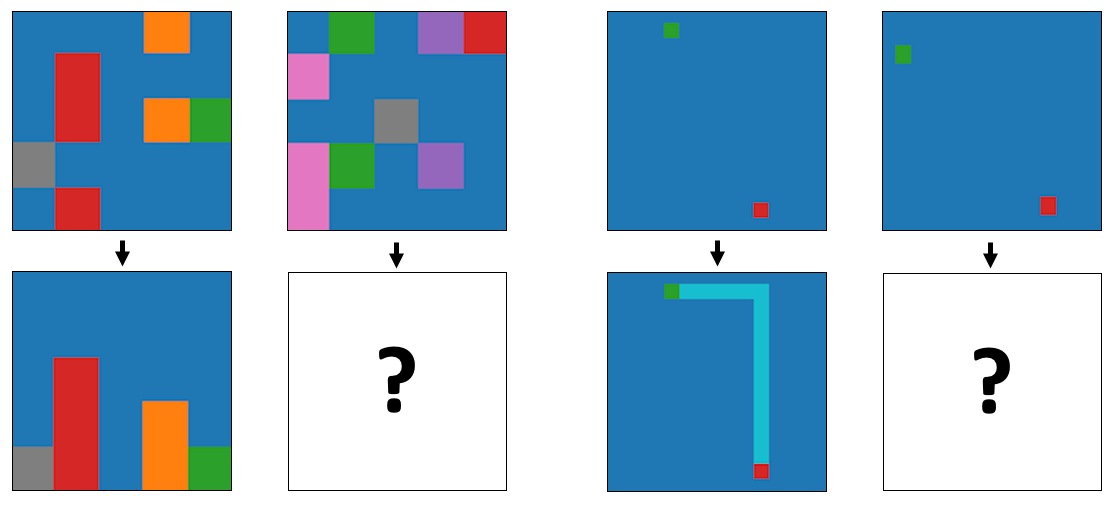}
        \caption{BEN does not solve either one of these two tasks, unlike the state-of-the-art Kaggle agent which has handcrafted primitives, one to 'simulate gravity' and another primitive which solves shortest path problems.}
        \label{fig:ARC_1e0a9b12_a2fd1cf0}
     \end{subfigure}
     \hfill
     \begin{subfigure}{0.49\linewidth}
        \centering
        \includegraphics[height=28mm]{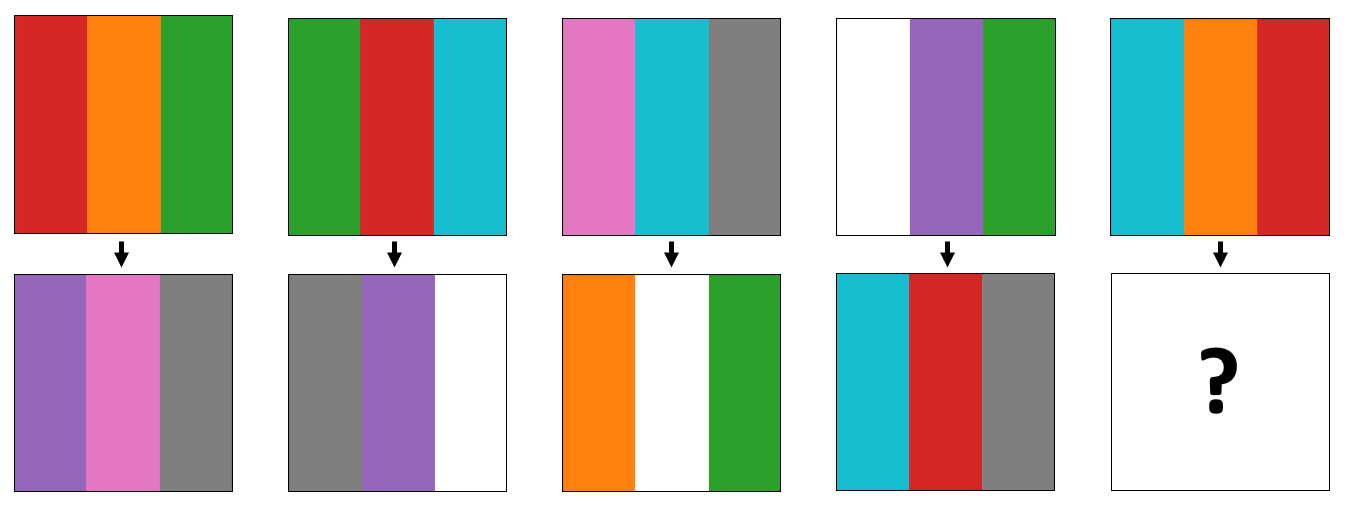}
        \caption{BEN executes a total of 8 synthesis loops to retrieve the 8 binary color swap transformation that make up the solution program, instead of synthesizing 8 binary color swap operations in a single deep program search.}
        \label{fig:ARC_0d3d703e}
     \end{subfigure}
        \caption{Performance on example visual reasoning tasks taken from ARC.}
\end{figure}

\paragraph{Knowledge Priors.}
We now take a look at specific task examples and work towards an ablation analysis of BEN's DA\&C framework in order to investigate which part of its performance is to due to better search versus only a refined domain-specific grammar.
Most of the tasks that BEN does not solve are due to either missing transformation primitives or control flow logic in its current grammar $\mathcal{G}_{ARC}$. The two tasks in \Cref{fig:ARC_1e0a9b12_a2fd1cf0} are examples of that:  
The first task requires a solution program that gravitates objects to the bottom of the frame. This is a complex motion program which has to enforce that objects closest to the bottom frame are translated first and that objects are only moved if their bottom neighboring pixel is empty. The second task is a version of a shortest path problem where the solution program is expected to find a trajectory which minimizes turns. Both of these tasks are only solved by the Kaggle agent \cite{wind_1st_2020} because it has access to handcrafted primitives in its prior transformation library, namely a 'gravity' operation and a 'shortest path' function which were custom engineered by its developer. Some of its 42 primitives only get used on one or a few tasks, whereas BEN only has access to 11 generic geometric transformations. In other words, the performance of the current state-of-the-art agent is in large parts due to hand-crafted, ad-hoc primitives, rather than better search. No combination of primitives in BEN's library (independent of the final size of a possible solution program) can reason about the path between two objects unless that path is a straight line. To make the comparison fairer, we conservatively reduce the agent's primitives to the ones that semantically match those in the domain grammar $\mathcal{G}_{ARC}$ of BEN (leaving 30 primitives). The matching was performed manually by studying the implementation of the primitives in the Kaggle agent. 
In this case, the performance of the Kaggle agent drops from 47\% to less than 25\%. That is, BEN solves a comparable amount of tasks with only a third of the primitives which is possible only because BEN decomposes tasks into recurring components which leads to simpler transformations with a more compact search space.
BEN's search space is more compact because a portion of the 30 primitives in the Kaggle agent serve as segmentation (e.g. filter by color) and selector functions (e.g. pick largest bitmap). In DA\&C, segmentation and selector primitives are part of the divide phase and the concept learning. They don't influence the size of the program search space in the main synthesis loop. In fact, selector primitives are not even searched but are entirely inferred from components by induction. Because the synthesis loop in the Kaggle agent is confounded with segmentation and selector primitives, it has to search through orders of magnitudes more candidate programs to achieve comparable expressiveness to that of the main synthesis loop in BEN augmented with a standalone segmentation and concept learning phase. Another related reason why DA\&C leads to a more compact search space are the combinations of segmentation and selector primitives in the DSL of the Kaggle agent. For instance, it includes a 'cutPickMax' primitive which combines the 'cut' and 'pickMax' primitives into a standalone function, presumably to make it available at lower search depths. BEN learns programs with the same semantics but offloads the 'pickMax' to the inductive concept learner which it can also apply to any other geometric primitive besides 'cut' with no impact on the search space within the main synthesis loop.
For the Kaggle agent, the computational challenge of synthesizing complex programs in a high-dimenional domain was overcome by hand-crafted sub programs which outsourced the cognitive effort to the developer rather than the program synthesis. We instead use ARC to demonstrate progress on the synthesis of complex programs from low-level primitives.
BEN uniquely solves 8\% of the data set on which the Kaggle agent fails even with its original transformation library of 42 primitives.

By splitting a deep program search across a linear number of much shallower synthesis searches, BEN is able to solve tasks which were previously out of reach for state-of-the-art agents on ARC. For example, the task in \Cref{fig:ARC_0d3d703e} requires an agent to learn a program that consists of 8 binary color swap operations. If an agent were to learn all eight operations in a single program search, the required search depth during synthesis is also eight, too deep for most grammars to still be exhaustively searchable. However, the color swap operations can be quickly learned from individual object tuples. In this case, the maximum search depth is reduced to one and the same search space is traversed eight times. BEN rapidly finds solutions to this and similar tasks.

\begin{figure}[b]
     \centering
     \begin{subfigure}{0.48\linewidth}
        \centering
        \includegraphics[scale=0.175]{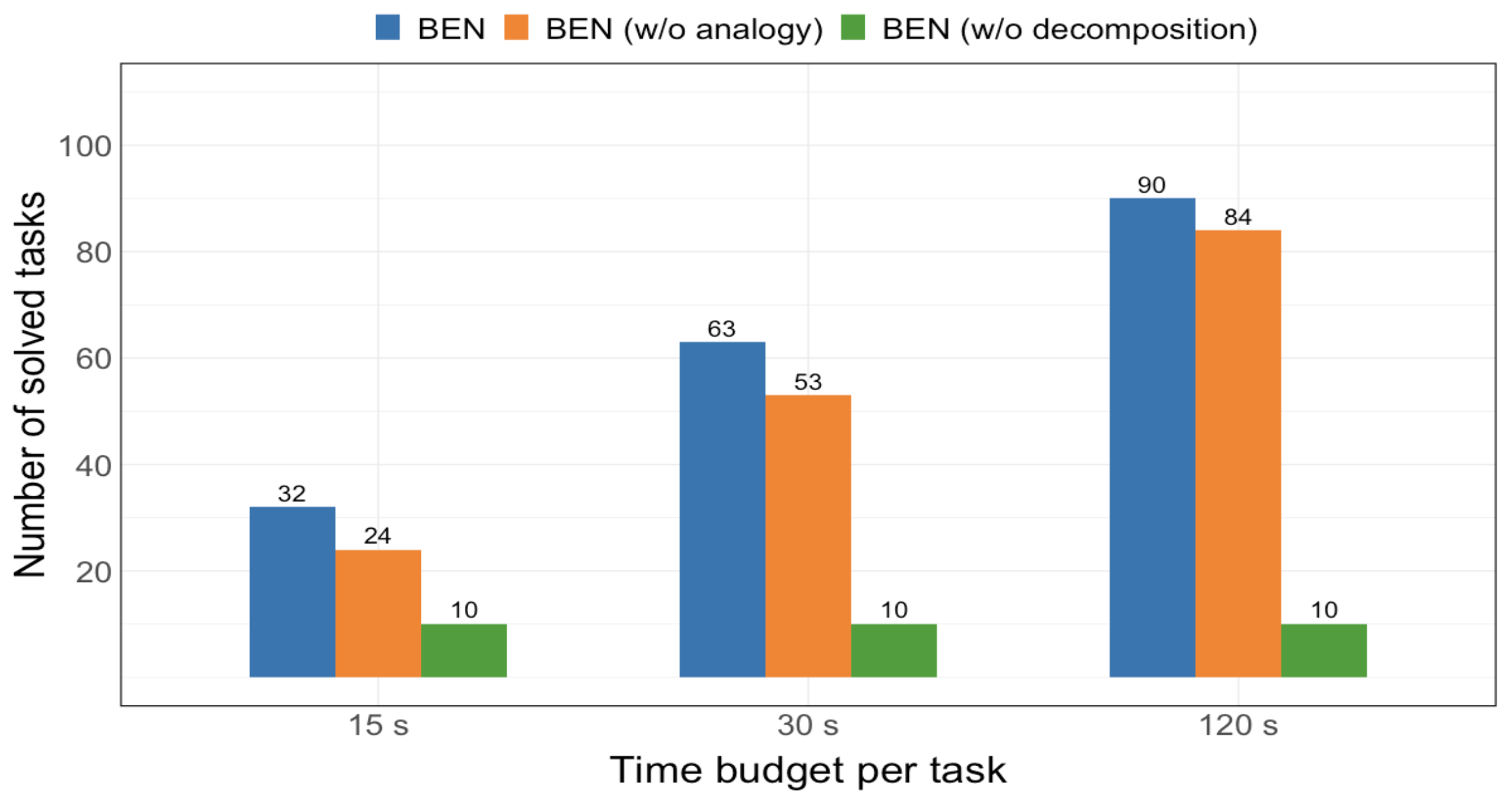}
        \caption{Performance of BEN's ablated baselines.}
        \label{fig:exp2_baselines}
     \end{subfigure}
     \hfill
     \begin{subfigure}{0.48\linewidth}
        \centering
        \includegraphics[scale=0.19]{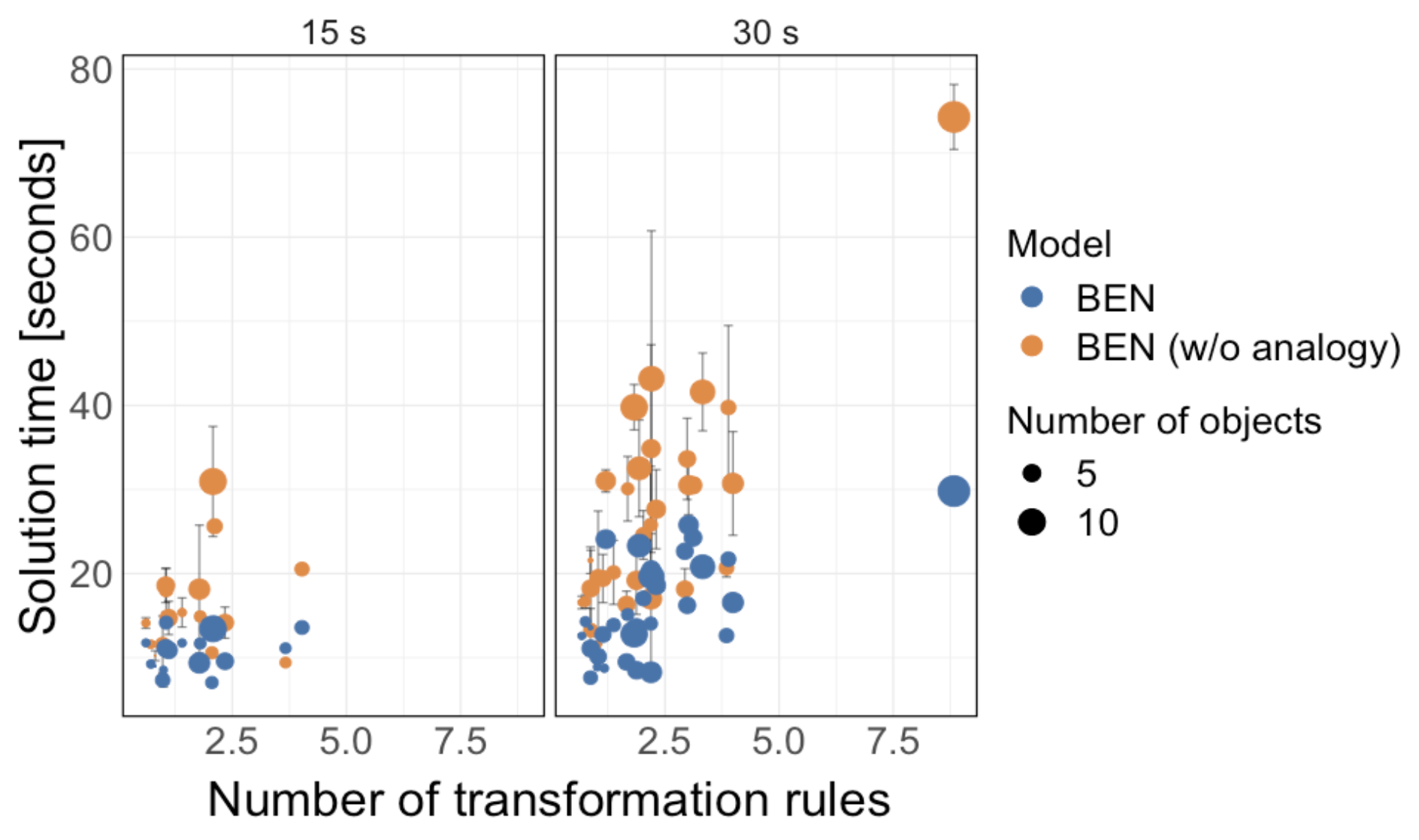}
        \caption{Solution times per program size.}
        \label{fig:exp2_solutiontimes}
     \end{subfigure}
        \caption{Performance benchmark on the abstract visual reasoning data sets from ARC. We report first found solutions for all ablations. (Time budgets in \Cref{fig:exp2_solutiontimes} are only enforced for BEN.)}
\end{figure}

\subsubsection{Ablation Analysis}
\label{subsec:baselines}
We compare to two ablated versions of BEN: one without analogical reasoning which performs synthesis on random object tuples and iterates trough them until it has discovered enough transformation programs to solve the task. The other without segmentation at all performing synthesis on the input/output images directly.

The results in \Cref{fig:exp2_baselines} show that analogical reasoning helps BEN (blue bars) solve more tasks in less time.
In the restrictive case of a 15s time budget per task, BEN solves roughly 33\% more tasks than its ablated baseline without analogical reasoning. The performance lead decreases to 7\% for a 2min time budget as the random search eventually covers a larger fraction of the total search space. Without segmentation, less than 12\% of tasks are being solved within 2min.
We verified that each of the tasks solved by either of the two baselines is also solved by BEN.

BEN has a average solution runtime of 15 seconds per task which is significantly lower than that of the random baseline, with a mean of 20 seconds, evaluated using a Wilcoxon signed-rank test on paired samples \textit{Z}=160, \textit{p}$<$.001 for a time budget of 30s per task (\Cref{fig:exp2_solutiontimes}).
The solution times include the time spent on segmentation and the time needed to identify meaningful object correspondences which shows that the additional computational effort of SME (structural alignment of the input/output examples) is limited and easily compensated for in the overall DA\&C algorithm.

\begin{table}[t]
\scriptsize
\centering
    \begin{tabular}{l | l || c c}
    \toprule
    \multicolumn{4}{c}{}\\
    Component & Ablation & 30 s & 120 s \\  \hline
    BEN & & 63 & 90 \\
    \hline
    \multirow{ 2}{*}{Divide} & w/o multicolor objects & -24\% & -33\% \\
     & w/o diagonal neighbors & -5\% & -22\% \\ \hline
    \multirow{ 3}{*}{Conquer: synthesis search} & depth $\leq$ 2 & -37\% & -50\%  \\
     & depth $\leq$ 3 & +25\% & -3\% \\
     & +25\% nonsense primitives in $\mathcal{G}_{transform}$ & -22\% & $\pm$0\% \\  \hline
    \multirow{ 3}{*}{Conquer: concept learner} & primary features & -34\% & -27\% \\
     & derived features & -24\% & -10\% \\
     & +50\% nonsense attributes in $\mathcal{G}_{concept}$ & $\pm$0\% & $\pm$0\% \\
    \hline
    \bottomrule
    \end{tabular}
\caption{An ablation analysis of BEN's core architecture design and domain specific knowledge priors shows that the contents of its transformation grammar $\mathcal{G}_{transform}$ and of its component attributes $\mathcal{G}_{concept}$ are especially robust to nonsense information.}
\label{tab:ablationanalysis}
\end{table}

The solution times in BEN remain low even for crowded scenes with a large number of decomposed objects ($\beta$ = 1.25, \textit{p}$<$.001), whereas the random baseline without analogical reasoning does not scale to larger scenes ($\beta$ = 3.81, \textit{p}$<$.001) consistent with previous observations on string transformation tasks \textbf{Q2}.
The program size in \Cref{fig:exp2_solutiontimes} refers to the number of transformation programs in the first found solution program where each of the transformation programs consists of multiple primitives and variables. The number of transformation programs is a lower bound on the number of synthesis searches that BEN needs to perform. Depending on the order in which correspondences are selected, more synthesis searches will be necessary. On average, BEN runs 4.7 synthesis searches on a task with an average number of correspondences of 16.4 where the first found solution program contains an average of 2.5 transformations. This means BEN explores about twice as many correspondences as would be minimally required to solve a task but only a fourth of all possible correspondences. This explains why program size is not a significant predictor of task solution times \textbf{Q3}. Instead, interaction of analogical program synthesis and the number of objects within a scene is a significant predictor of task solution times (F(3,176)=24.6, \textit{p}$<$.001). This confirms that the guidance from analogical synthesis is especially helpful for tasks that require learning transformations over large scenes with many components.

In order to investigate the impact of the DA\&C design on BEN's performance, we systematically manipulate its search as well as its domain grammars and report the ablations in \Cref{tab:ablationanalysis}.
Most notably, adding as much as 50\% additional nonsense attributes to component encodings $\mathcal{G}_{concept}$ or 25\% additional nonsense primitives to the transformation grammar $\mathcal{G}_{transform}$ only has a small effect on BEN's performance. In fact, adding more nonsense features neither caused BEN to miss tasks nor did it increase solutions times. This is because we learn concepts bottom-up on successful transformation programs instead of searching them as logical formulas in a domain grammar which is what ARGA does. Additional transformation primitives prolong the time of a single synthesis search and, therefore, cause BEN to timeout on 22\% of tasks it had previously solved within a challenging 30 s time budget per task. Its performance under a time budget of 120 s remains unaffected. We present this as evidence that the domain-specific knowledge provided to BEN has not been custom engineered and its performance is mainly due to intelligent hierarchical search in the DA\&C framework.
Reducing the depth of the synthesis search generally leads to decreased performance. On small time budgets, however, a slight reduction in search depth can have the opposite effect when the majority of sub programs (needed for a solution program) can be retrieved from a shallow synthesis search. As for the decomposition language, its expressiveness (e.g., diagonal neighborhood, multicolor segmentation) directly impacts the agent's task performance.

\begin{figure}[t]
     \centering
     \begin{subfigure}{0.49\linewidth}
        \centering
        \includegraphics[scale=0.4]{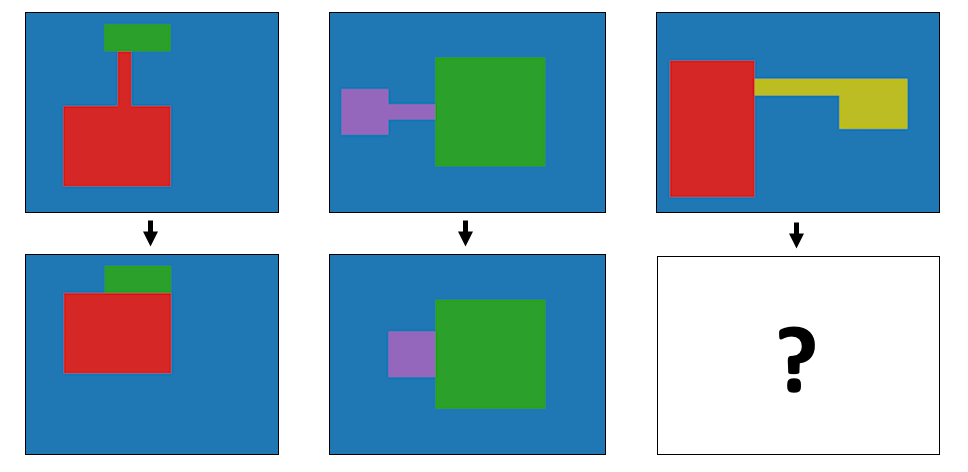}
        \caption{BEN learns a program which moves both objects together, thereby compressing the shape of the elongated object.}
        \label{fig:ARC_98cf29f8}
     \end{subfigure}
     \hfill
     \begin{subfigure}{0.49\linewidth}
        \centering
        \includegraphics[scale=0.25]{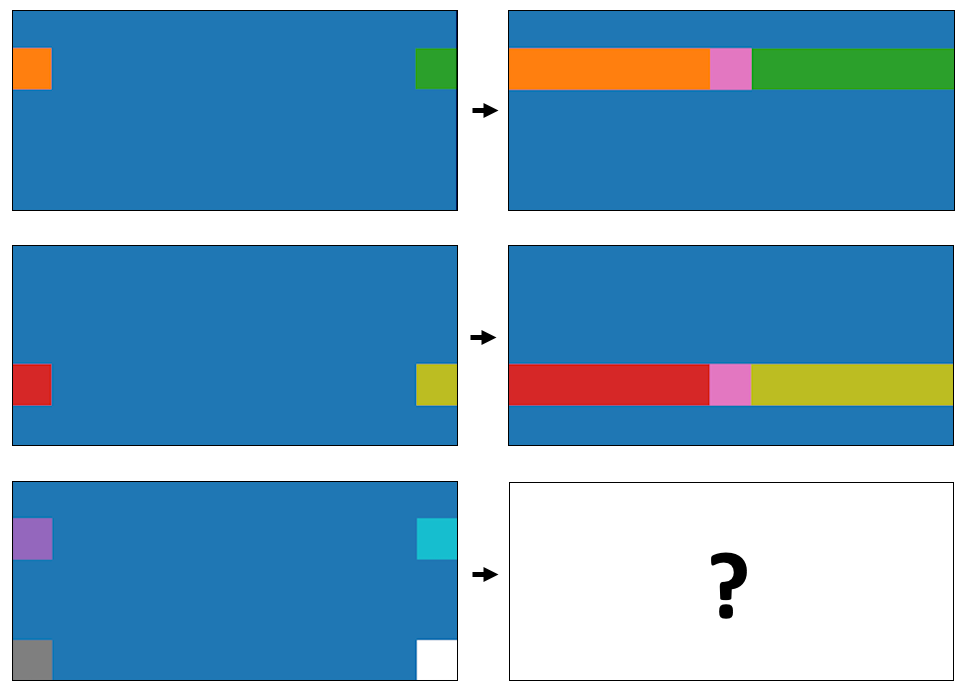}
        \caption{BEN also invents new objects in the output and can take into account information from the input such as the x-coordinate at which an object should be created.}
        \label{fig:ARC_29c11459}
     \end{subfigure}
        \caption{BEN also solves ARC tasks which involve n-1 mappings or require the generation of new components in the output.}
\end{figure}

\paragraph{1-1 Correspondences.}
On some of the tasks, it is counter-intuitive how the reliance on pairwise correspondences is improving rather than restricting search. We find that it is a viable heuristic to guide the program search even in cases where there exist no obvious 1-1 mappings between input and output components. For instance, n-1 mappings frequently occur in visual reasoning tasks in which information from multiple input components is pooled (e.g. shape and color from two different components) to reconstruct a component in the output. Another example of n-1 mappings are directed movements towards other objects (\Cref{fig:ARC_98cf29f8}). In this examples, BEN still finds the correct transformation starting from any of the 1-1 correspondences within the n-1 mapping because its transformation grammar is able to refer to knowledge from other input components (\Cref{fig:Gdomain_ARC}). We have introduced the 'REFERENCE' non-terminal in $\mathcal{G}_{transform}$ to make this explicit. The references in this paper are hard coded ('largest object', 'other object' ...) but this does not need to be the case. Similar to how DA\&C uses constraint solving to learn a logic formula over input components that make use of the same transformation program, constraint solving could be used to learn a logic formula over 'referred components' within transformation programs. BEN also solves tasks which require the invention of new objects in the output such as the example in \Cref{fig:ARC_29c11459}. The correspondence which BEN leverages to synthesize the pink pixel could be with any of the two pixels in the input.
If the alignment produces 1-1 correspondences which don't yield a meaningful transformation program, those will never cause BEN to miss a task which it had otherwise solved, because the alignment is only used as a heuristic on the order in which to search through the program space. In the worst case, BEN explores the Cartesian product of all correspondences between the segmented input and segmented output. We emphasize that DAC is no universal strategy to all synthesis problems. It is ideally suited for those with compositional structure. If task examples have no inherent compositional structure, BEN degrades to program synthesis search on the unsegmented input/output examples.

\paragraph{Importance of the family of decomposition functions $\delta$.}
In order to analyze the impact of the prior knowledge of $\mathcal{G}_{decomp}$ on BEN's performance, we conducted an exploratory study on the solution strategies of human task solvers (\textit{N} = 4) (\textit{publication reference anonymized for blind review})\ignore{\cite{witt_grammar-based_2021}}: Participants were presented with ARC tasks on a web interface which asked them to reconstruct the missing output for each test case and to highlight individual objects in each image tuple $(I_l, R_l) \in \mathcal{Q}$. We used the reconstruction accuracy on test cases to control if participants understood a task. Only the image segmentations from successful task solvers were considered in the subsequent analysis.


We find that the segmentations produced by BEN perfectly match those of human task solvers for roughly 38\% of the tasks.
Whenever a segmentation deviates from those of human participants, BEN is less likely to solve a task. The introductory ARC task in \Cref{fig:exampletask_ARC} is an example of the contrary: BEN finds an alternative segmentation in which directly neighboring light-blue and green pixels are merged into the same object. The output is colored light-blue whenever there exists a multicolored object in the input and its width is larger than 3. This solution program does not fully capture the semantics of the task but is sufficient to solve all test cases.
In order to estimate the percentage of tasks that BEN fails to solve due to insufficient segmentation, we randomly choose 50 failed tasks and directly execute BEN on segmentations produced by human task solvers. 
BEN then solves 16\% of these previously unsolved tasks which suggests that missing transformation primitives and object features are the main bottleneck instead of the current segmentation grammar.

\section{Discussion And Future Work}
Humans effortlessly induce large programs which generalize well to previously unseen test cases, even in a few-shot learning setting, as in ARC or the real world string transformations data set. 
Our work suggests that program synthesis can exploit the compositional nature of structured domains to guide the search for well-generalizing programs in vast language spaces and in high dimensional domains.
That is done by decomposing the problem of learning a single nested program into a two stage process: first, we find a segmentation of examples into meaningful components and then perform multiple program synthesis tasks on component tuples in the input/output.
Second, separating the problem of searching for component-specific transformations from the task of learning the contexts in which they apply (the concept definition) leads to better generalizing programs.
We implemented the DA\&C paradigm in our agent BEN using top-down enumerative search as a synthesis technique. Future work that evaluates the integration with other advanced synthesis techniques appears promising. DA\&C is a program synthesis framework which exploits the compositional structure in task examples and otherwise degrades to whichever synthesis technique it uses in its main loop.

Although BEN solves more string transformation tasks than state-of-the-art ILP baselines and a fair share of highly heterogeneous visual reasoning tasks, it does not yet match the performance of an average human task solver. Humans seem to use additional strategies that deviate from DA\&C and traditional search-based synthesis in important ways.
For example, humans make efficient use of context switching, where the synthesis has access to only those object features and transformation primitives which appear important to the task at hand. This reduces the load of having to process many different encodings all at once as well as limits the search space. Neurally-guided search could be a computational means to this which would require an additional learning component or the neural implementation of one of the DA\&C subroutines.

It is well established in the cognitive sciences that higher-order functions in human cognition, such as abstract reasoning over programs, influence the grouping of perceptual units and the encoding of their features. In the future, we will explore how the decomposition phase could be conditioned on the progress of the synthesis to group together what is likely to act as a single functional unit in the solution program.

Finally, the assumption of independent object transformations is a simplification. Humans are readily able to find a suitable sequential order of object transformations for example to make use of overpainting. The methods proposed in this paper do not rely on independent object transformations per se and, hence, there is the need to investigate ways of composing programs with complex dependencies between their transformations.

\bibliography{jair2023}

\begin{thebibliography}{}

\bibitem[\protect\BCAY{Alur, Radhakrishna,\ \BBA\ Udupa}{Alur et~al.}{2017}]{alur_scaling_2017}
Alur, R., Radhakrishna, A., \BBA\ Udupa, A. \BBOP2017\BBCP.
\newblock \BBOQ Scaling enumerative program synthesis via divide and conquer\BBCQ\
\newblock In {\Bem Tools and {Algorithms} for the {Construction} and {Analysis} of {Systems} ({TACAS})}.

\bibitem[\protect\BCAY{Alur, Singh, Fisman,\ \BBA\ Solar-Lezama}{Alur et~al.}{2018}]{alur_search-based_2018}
Alur, R., Singh, R., Fisman, D., \BBA\ Solar-Lezama, A. \BBOP2018\BBCP.
\newblock \BBOQ Search-based program synthesis\BBCQ\
\newblock {\Bem Communications of the ACM}, {\Bem 61\/}(12), 84--93.

\bibitem[\protect\BCAY{Alur, Černý,\ \BBA\ Radhakrishna}{Alur et~al.}{2015}]{alur_synthesis_2015}
Alur, R., Černý, P., \BBA\ Radhakrishna, A. \BBOP2015\BBCP.
\newblock \BBOQ Synthesis through unification\BBCQ\
\newblock In Kroening, D.\BBACOMMA\  \BBA\ Păsăreanu, C.~S.\BEDS, {\Bem Computer {Aided} {Verification}}, \BPGS\ 163--179, Cham. Springer International Publishing.

\bibitem[\protect\BCAY{Chollet}{Chollet}{2019}]{chollet_measure_2019}
Chollet, F. \BBOP2019\BBCP.
\newblock \BBOQ On the {Measure} of {Intelligence}\BBCQ\
\newblock ArXiv. https://doi.org/10.48550/arXiv.1911.01547.

\bibitem[\protect\BCAY{Chollet}{Chollet}{2020}]{chollet_abstraction_2020}
Chollet, F. \BBOP2020\BBCP.
\newblock \BBOQ The {Abstraction} and {Reasoning} {Corpus} ({ARC})\BBCQ\
\newblock GitHub. https://github.com/fchollet/ARC.

\bibitem[\protect\BCAY{Cropper}{Cropper}{2022}]{cropper_learning_2022}
Cropper, A. \BBOP2022\BBCP.
\newblock \BBOQ Learning logic programs through divide, constrain, and conquer\BBCQ\
\newblock In {\Bem 36th {Conference} on {Artificial} {Intelligence}, {AAAI} 2022, {Virtual} {Event}, {February} 22 - {March} 1}, \BPGS\ 6446--6453. AAAI Press.

\bibitem[\protect\BCAY{Cropper\ \BBA\ Dumančić}{Cropper\ \BBA\ Dumančić}{2020}]{cropper_learning_2020}
Cropper, A.\BBACOMMA\  \BBA\ Dumančić, S. \BBOP2020\BBCP.
\newblock \BBOQ Learning large logic programs by going beyond entailment\BBCQ\
\newblock In Bessiere, C.\BED, {\Bem Proceedings of the {Twenty}-{Ninth} {International} {Joint} {Conference} on {Artificial} {Intelligence}, {IJCAI} 2020}, \BPGS\ 2073--2079.

\bibitem[\protect\BCAY{Cropper\ \BBA\ Dumančić}{Cropper\ \BBA\ Dumančić}{2022}]{cropper_inductive_2022}
Cropper, A.\BBACOMMA\  \BBA\ Dumančić, S. \BBOP2022\BBCP.
\newblock \BBOQ Inductive {Logic} {Programming} {At} 30: {A} {New} {Introduction}\BBCQ\
\newblock {\Bem Journal of Artificial Intelligence Research}, {\Bem 74}, 765--850.

\bibitem[\protect\BCAY{Cropper, Morel,\ \BBA\ Muggleton}{Cropper et~al.}{2020}]{cropper_learning_2020-1}
Cropper, A., Morel, R., \BBA\ Muggleton, S.~H. \BBOP2020\BBCP.
\newblock \BBOQ Learning higher-order programs through predicate invention\BBCQ\
\newblock {\Bem Proceedings of the AAAI Conference on Artificial Intelligence}, {\Bem 34\/}(09), 13655--13658.

\bibitem[\protect\BCAY{Cropper\ \BBA\ Muggleton}{Cropper\ \BBA\ Muggleton}{2016}]{cropper_metagol_2016}
Cropper, A.\BBACOMMA\  \BBA\ Muggleton, S.~H. \BBOP2016\BBCP.
\newblock \BBOQ Metagol system\BBCQ\
\newblock GitHub. https://github.com/metagol/metagol.

\bibitem[\protect\BCAY{de~Miquel, Corominas,\ \BBA\ Ariyasu}{de~Miquel et~al.}{2020}]{de_miquel_2nd_2020}
de~Miquel, A., Corominas, R.~G., \BBA\ Ariyasu, Y. \BBOP2020\BBCP.
\newblock \BBOQ 2nd place solution {ARC} {Kaggle} competition\BBCQ\
\newblock GitHub. https://github.com/alejandrodemiquel/ARC\_Kaggle.

\bibitem[\protect\BCAY{Dumančić, Guns,\ \BBA\ Cropper}{Dumančić et~al.}{2021}]{dumancic_knowledge_2021}
Dumančić, S., Guns, T., \BBA\ Cropper, A. \BBOP2021\BBCP.
\newblock \BBOQ Knowledge refactoring for inductive program synthesis\BBCQ\
\newblock In {\Bem 35th {AAAI} {Conference} on {Artificial} {Intelligence}, {AAAI} 2021, {Virtual} {Event}, {February} 2-9}, \BPGS\ 7271--7278. AAAI Press.

\bibitem[\protect\BCAY{Ellis, Nye, Pu, Sosa, Tenenbaum,\ \BBA\ Solar-Lezama}{Ellis et~al.}{2019}]{ellis_write_2019}
Ellis, K., Nye, M.~I., Pu, Y., Sosa, F., Tenenbaum, J., \BBA\ Solar-Lezama, A. \BBOP2019\BBCP.
\newblock \BBOQ Write, execute, assess: {Program} synthesis with a {REPL}\BBCQ\
\newblock In Wallach, H.~M., Larochelle, H., Beygelzimer, A., d'Alché Buc, F., Fox, E.~B., \BBA\ Garnett, R.\BEDS, {\Bem Advances in {Neural} {Information} {Processing} {Systems}: {Annual} {Conference} on {Neural} {Information} {Processing} {Systems} 2019, {NeurIPS} 2019, {December} 8-14}, \lowercase{\BVOL}~32, \BPGS\ 9165--9174, Vancouver, BC, Canada. Curran Associates Inc.

\bibitem[\protect\BCAY{Ellis, Solar-Lezama,\ \BBA\ Tenenbaum}{Ellis et~al.}{2015}]{ellis_unsupervised_2015}
Ellis, K., Solar-Lezama, A., \BBA\ Tenenbaum, J. \BBOP2015\BBCP.
\newblock \BBOQ Unsupervised learning by program synthesis\BBCQ\
\newblock In Cortes, C., Lawrence, N., Lee, D., Sugiyama, M., \BBA\ Garnett, R.\BEDS, {\Bem Advances in {Neural} {Information} {Processing} {Systems}: {Annual} {Conference} on {Neural} {Information} {Processing} {Systems} 2015, {NeurIPS} 2015, {December} 7-10}, \lowercase{\BVOL}~28, Montreal, QC, Canada. Curran Associates, Inc.

\bibitem[\protect\BCAY{Evans}{Evans}{1964}]{evans_heuristic_1964}
Evans, T.~G. \BBOP1964\BBCP.
\newblock \BBOQ A heuristic program to solve geometric-analogy problems\BBCQ\
\newblock In {\Bem Proceedings of the {Spring} {Joint} {Computer} {Conference}, {AFIPS} 1964, {April} 21-23}, \BPGS\ 327--338, New York, NY, USA. Association for Computing Machinery.

\bibitem[\protect\BCAY{Falkenhainer, Forbus,\ \BBA\ Gentner}{Falkenhainer et~al.}{1989}]{falkenhainer_structure-mapping_1989}
Falkenhainer, B., Forbus, K.~D., \BBA\ Gentner, D. \BBOP1989\BBCP.
\newblock \BBOQ The structure-mapping engine: {Algorithm} and examples\BBCQ\
\newblock {\Bem Artificial Intelligence}, {\Bem 41\/}(1), 1--63.

\bibitem[\protect\BCAY{Gentner}{Gentner}{1983}]{gentner_structure-mapping_1983}
Gentner, D. \BBOP1983\BBCP.
\newblock \BBOQ Structure-mapping: a theoretical framework for analogy\BBCQ\
\newblock {\Bem Cognitive Science}, {\Bem 7\/}(2), 155--170.

\bibitem[\protect\BCAY{Gulwani}{Gulwani}{2011}]{gulwani_automating_2011}
Gulwani, S. \BBOP2011\BBCP.
\newblock \BBOQ Automating string processing in spreadsheets using input-output examples\BBCQ\
\newblock In {\Bem Principles of {Programming} {Languages}, {POPL} 2011, {January} 26-28}, Austin, Texas, USA.

\bibitem[\protect\BCAY{Gulwani\ \BBA\ Jain}{Gulwani\ \BBA\ Jain}{2017}]{gulwani_programming_2017}
Gulwani, S.\BBACOMMA\  \BBA\ Jain, P. \BBOP2017\BBCP.
\newblock \BBOQ Programming by examples: {PL} meets {ML}\BBCQ\
\newblock In {\Bem Asian {Symposium} on {Programming} {Languages} and {Systems}, {APLAS} 2017, {November} 27-29}, Suzhou, China. Springer.

\bibitem[\protect\BCAY{Guns, Nijssen,\ \BBA\ De~Raedt}{Guns et~al.}{2013}]{guns_k-pattern_2013}
Guns, T., Nijssen, S., \BBA\ De~Raedt, L. \BBOP2013\BBCP.
\newblock \BBOQ k-{Pattern} set mining under constraints\BBCQ\
\newblock {\Bem IEEE Transactions on Knowledge and Data Engineering}, {\Bem 25\/}(2), 402--418.

\bibitem[\protect\BCAY{Henderson\ \BBA\ Muggleton}{Henderson\ \BBA\ Muggleton}{2014}]{henderson_automatic_2014}
Henderson, R.\BBACOMMA\  \BBA\ Muggleton, S. \BBOP2014\BBCP.
\newblock \BBOQ Automatic invention of functional abstractions\BBCQ\
\newblock In {\Bem Latest {Advances} in {Inductive} {Logic} {Programming}}, \BPGS\ 217--224. Imperial College Press, London, UK.

\bibitem[\protect\BCAY{Hofstadter}{Hofstadter}{2001}]{hofstadter_analogy_2001}
Hofstadter, D. \BBOP2001\BBCP.
\newblock \BBOQ Analogy as the core of cognition\BBCQ\
\newblock In Gentner, D., Holyoak, K., \BBA\ Kokinov, B.\BEDS, {\Bem The {Analogical} {Mind}: {Perpectives} from {Cognitive} {Science}}, \BPGS\ 499--538. MIT Press, Cambridge, MA, USA.

\bibitem[\protect\BCAY{Johnson, Vong, Lake,\ \BBA\ Gureckis}{Johnson et~al.}{2021}]{johnson_fast_2021}
Johnson, A., Vong, W.~K., Lake, B.~M., \BBA\ Gureckis, T.~M. \BBOP2021\BBCP.
\newblock \BBOQ Fast and flexible: {Human} program induction in abstract reasoning tasks\BBCQ\
\newblock ArXiv. https://doi.org/10.48550/arXiv.2103.05823.

\bibitem[\protect\BCAY{Lin, Dechter, Ellis, Tenenbaum,\ \BBA\ Muggleton}{Lin et~al.}{2014}]{lin_bias_2014}
Lin, D., Dechter, E., Ellis, K., Tenenbaum, J., \BBA\ Muggleton, S. \BBOP2014\BBCP.
\newblock \BBOQ Bias reformulation for one-shot function induction\BBCQ\
\newblock In {\Bem Proceedings of the {Twenty}-{First} {European} {Conference} on {Artificial} {Intelligence}, {ECAI} 2014}, \BPGS\ 525--530, Prague, Czech Republic.

\bibitem[\protect\BCAY{Lovett\ \BBA\ Forbus}{Lovett\ \BBA\ Forbus}{2017}]{lovett_modeling_2017}
Lovett, A.\BBACOMMA\  \BBA\ Forbus, K.~D. \BBOP2017\BBCP.
\newblock \BBOQ Modeling visual problem solving as analogical reasoning\BBCQ\
\newblock {\Bem Psychological Review}, {\Bem 124\/}(1), 60--90.

\bibitem[\protect\BCAY{Mitchell}{Mitchell}{1993}]{mitchell_analogy-making_1993}
Mitchell, M. \BBOP1993\BBCP.
\newblock {\Bem Analogy-making as perception: {A} computer model}.
\newblock MIT Press, Cambridge, MA, USA.

\bibitem[\protect\BCAY{Mitchell}{Mitchell}{2021}]{mitchell_abstraction_2021}
Mitchell, M. \BBOP2021\BBCP.
\newblock \BBOQ Abstraction and analogy-making in artificial intelligence\BBCQ\
\newblock {\Bem Annals of the New York Academy of Sciences}, {\Bem 1505\/}(1), 79--101.

\bibitem[\protect\BCAY{Muggleton, Lin,\ \BBA\ Tamaddoni-Nezhad}{Muggleton et~al.}{2015}]{muggleton_meta-interpretive_2015}
Muggleton, S.~H., Lin, D., \BBA\ Tamaddoni-Nezhad, A. \BBOP2015\BBCP.
\newblock \BBOQ Meta-interpretive learning of higher-order dyadic datalog: {Predicate} invention revisited\BBCQ\
\newblock {\Bem Machine Learning}, {\Bem 100\/}(1), 49--73.

\bibitem[\protect\BCAY{Neider, Saha,\ \BBA\ Madhusudan}{Neider et~al.}{2016}]{neider_synthesizing_2016}
Neider, D., Saha, S., \BBA\ Madhusudan, P. \BBOP2016\BBCP.
\newblock \BBOQ Synthesizing piece-wise functions by learning classifiers\BBCQ\
\newblock In {\Bem Proceedings of the 22nd {International} {Conference} on {Tools} and {Algorithms} for the {Construction} and {Analysis} of {Systems}, {TACAS} 2016}, \lowercase{\BVOL}\ 9636, \BPGS\ 186--203, Berlin, Germany. Springer-Verlag.

\bibitem[\protect\BCAY{Nye, Pu, Bowers, Andreas, Tenenbaum,\ \BBA\ Solar-Lezama}{Nye et~al.}{2021}]{nye_representing_2021}
Nye, M.~I., Pu, Y., Bowers, M., Andreas, J., Tenenbaum, J.~B., \BBA\ Solar-Lezama, A. \BBOP2021\BBCP.
\newblock \BBOQ Representing partial programs with blended abstract semantics\BBCQ\
\newblock In {\Bem 9th {International} {Conference} on {Learning} {Representations}, {ICLR} 2021, {Virtual} {Event}, {May} 3-7}.

\bibitem[\protect\BCAY{Raza\ \BBA\ Gulwani}{Raza\ \BBA\ Gulwani}{2017}]{raza_automated_2017}
Raza, M.\BBACOMMA\  \BBA\ Gulwani, S. \BBOP2017\BBCP.
\newblock \BBOQ Automated data extraction using predictive program synthesis\BBCQ\
\newblock In {\Bem 31st {Conference} on {Artificial} {Intelligence}, {AAAI} 2017, {February} 4-9}, San Francisco, CA, USA. Association for the Advancement of Artificial Intelligence.

\bibitem[\protect\BCAY{Snow, Kyllonen,\ \BBA\ Marshalek}{Snow et~al.}{1984}]{snow_topography_1984}
Snow, R.~E., Kyllonen, P.~C., \BBA\ Marshalek, B. \BBOP1984\BBCP.
\newblock \BBOQ The topography of ability and learning correlations\BBCQ\
\newblock In Sternberg, R.~J.\BED, {\Bem Advances in the psychology of human intelligence}, \BPGS\ 47--103. Erlbaum, Hillsdale, NJ, USA.

\bibitem[\protect\BCAY{{Sumit Gulwani}}{{Sumit Gulwani}}{2023}]{sumit_gulwani_microsoft_2023}
{Sumit Gulwani} \BBOP2023\BBCP.
\newblock \BBOQ Microsoft program synthesis using examples ({PROSE})\BBCQ\
\newblock GitHub. https://github.com/microsoft/prose.

\bibitem[\protect\BCAY{Valiant}{Valiant}{1985}]{valiant_learning_1985}
Valiant, L.~G. \BBOP1985\BBCP.
\newblock \BBOQ Learning disjunction of conjunctions\BBCQ\
\newblock In {\Bem Proceedings of the 9th {International} {Joint} {Conference} on {Artificial} {Intelligence}, {IJCAI} 1985}, \lowercase{\BVOL}~1, \BPGS\ 560--566, San Francisco, CA, USA. Morgan Kaufmann Publishers Inc.

\bibitem[\protect\BCAY{Wagemans, Elder, Kubovy, Palmer, Peterson, Singh,\ \BBA\ Heydt}{Wagemans et~al.}{2012}]{wagemans_century_2012}
Wagemans, J., Elder, J., Kubovy, M., Palmer, S., Peterson, M., Singh, M., \BBA\ Heydt, R. \BBOP2012\BBCP.
\newblock \BBOQ A century of gestalt psychology in visual perception: {I}. {Perceptual} grouping and figure-ground organization\BBCQ\
\newblock {\Bem Psychological Bulletin}, {\Bem 138\/}(6), 1172--1217.

\bibitem[\protect\BCAY{Wang, Cheung,\ \BBA\ Bodik}{Wang et~al.}{2017}]{wang_synthesizing_2017}
Wang, C., Cheung, A., \BBA\ Bodik, R. \BBOP2017\BBCP.
\newblock \BBOQ Synthesizing highly expressive {SQL} queries from input-output examples\BBCQ\
\newblock In {\Bem Proceedings of the 38th {Conference} on {Programming} {Language} {Design} and {Implementation}, {PLDI} 2017, {June} 18-23}, \BPGS\ 452--466, New York, NY, USA. Association for Computing Machinery.

\bibitem[\protect\BCAY{Wind}{Wind}{2020}]{wind_1st_2020}
Wind, J.~S. \BBOP2020\BBCP.
\newblock \BBOQ 1st place solution {ARC} {Kaggle} competition\BBCQ\
\newblock GitHub. https://github.com/top-quarks/ARC-solution.

\bibitem[\protect\BCAY{Xu, Khalil,\ \BBA\ Sanner}{Xu et~al.}{2022}]{xu_graphs_2022}
Xu, Y., Khalil, E.~B., \BBA\ Sanner, S. \BBOP2022\BBCP.
\newblock \BBOQ Graphs, {Constraints}, and {Search} for the {Abstraction} and {Reasoning} {Corpus}\BBCQ\
\newblock In {\Bem In 37th {AAAI} {Conference} on {Artificial} {Intelligence}, {AAAI} 2023}.

\end{thebibliography}
\bibliographystyle{theapa}

\end{document}